\documentclass[journal]{IEEEtran}
\usepackage{booktabs}
\usepackage{amssymb}
\usepackage{amsfonts}
\usepackage{bbding}
\usepackage{svg}
\usepackage{subfigure}
\usepackage{amsmath} 
\usepackage[numbers]{natbib}
\usepackage{url}

\ifCLASSOPTIONcompsoc
 \usepackage[caption=false,font=normalsize,labelfont=sf,textfont=sf]{subfig}
\else
 \usepackage[caption=false,font=footnotesize]{subfig}
\fi
\usepackage{dblfloatfix}
\usepackage{url}
\begin{document}

\def\UrlFont{\em}

\title{RS-Mamba for Large Remote Sensing Image Dense Prediction}

\author{Sijie~Zhao,
 Hao~Chen*,
 Xueliang~Zhang*,
 Pengfeng~Xiao,
 Lei Bai,
 and Wanli Ouyang
 \thanks{Sijie Zhao, Xueliang Zhang, and Pengfeng Xiao are with the Jiangsu Provincial Key Laboratory of Geographic
Information Science and Technology, Key Laboratory for Land Satellite Remote Sensing Applications of Ministry of Natural Resources, School of Geography and Ocean Science, Nanjing University, Nanjing 210023, China (e-mail: zsj@smail.nju.edu.cn; zxl@nju.edu.cn; xiaopf@nju.edu.cn).

Hao Chen, Lei Bai, and Wanli Ouyang is with the Shanghai Artificial Intelligence Laboratory, Shanghai 200000, China (e-mail: justchenhao@buaa.edu.cn, bailei@pjlab.org.cn, ouyangwanli@pjlab.org.cn).

Corresponding Author: Hao Chen and Xueliang Zhang.
}
}

\maketitle

\begin{abstract}
Context modeling is critical for remote sensing image dense prediction tasks. Nowadays, the growing size of very-high-resolution (VHR) remote sensing images poses challenges in effectively modeling context. While transformer-based models possess global modeling capabilities, they encounter computational challenges when applied to large VHR images due to their quadratic complexity. The conventional practice of cropping large images into smaller patches results in a notable loss of contextual information. To address these issues, we propose the Remote Sensing Mamba (RSM) for dense prediction tasks in large VHR remote sensing images. RSM is specifically designed to capture the global context of remote sensing images with linear complexity, facilitating the effective processing of large VHR images. Considering that the land covers in remote sensing images are distributed in arbitrary spatial directions due to characteristics of remote sensing over-head imaging, the RSM incorporates an omnidirectional selective scan module to globally model the context of images in multiple directions, capturing large spatial features from various directions. Extensive experiments on semantic segmentation and change detection tasks across various land covers demonstrate the effectiveness of the proposed RSM. We designed simple yet effective models based on RSM, achieving state-of-the-art performance on dense prediction tasks in VHR remote sensing images without fancy training strategies. Leveraging the linear complexity and global modeling capabilities, RSM achieves better efficiency and accuracy than transformer-based models on large remote sensing images. Interestingly, we also demonstrated that our model generally performs better with a larger image size on dense prediction tasks. Our code is available at \protect\url{https://github.com/walking-shadow/Official_Remote_Sensing_Mamba}.

% The spatial resolution of remote sensing images is becoming increasingly high, posing challenges in handling large very-high-resolution (VHR) remote sensing images for dense prediction tasks. Models based on convolutional neural networks are limited in their ability to model global features of remote sensing images due to local convolution operations. Despite their global modeling capabilities, transformer-based models face computational challenges with large VHR images due to their quadratic complexity. The common practice of cropping large images into smaller patches leads to a significant loss of contextual information. To address these issues, we propose the Remote Sensing Mamba (RSM) for dense prediction tasks in VHR remote sensing. RSM is designed to model global features of remote sensing images with linear complexity, enabling it to process large VHR images effectively. Given that the spatial features of remote sensing images are distributed in arbitrary spatial directions, RSM employs an omnidirectional selective scan module to globally model images in multiple directions, capturing large spatial features from various directions. Experiments on semantic segmentation and change detection tasks across various objects demonstrate the effectiveness of the RSM. With a simple model architecture and training approach, the RSM achieves state-of-the-art performance on dense prediction tasks of VHR remote sensing. The code for this work is available at \url{https://github.com/walking-shadow/Official_Remote_Sensing_Mamba}.

\end{abstract}


\IEEEpeerreviewmaketitle

\section{Introduction}

\IEEEPARstart{T}{he} advent of increasingly high spatial resolution in remote sensing images has marked a transformative period in the field, facilitating a deeper understanding and more nuanced analysis across a multitude of applications. These high-resolution images serve as pivotal resources in various domains, including urban planning~\cite{urban_planning}, agricultural management~\cite{agricultural_management}, environmental monitoring~\cite{environmental_monitoring}, and disaster response~\cite{disaster_response}. 

Due to the unprecedented availability of very-high-resolution (VHR) images, the field of remote sensing has undergone rapid expansion in recent years. VHR remote sensing images are characterized by rich contextual information, which is crucial for dense prediction tasks such as semantic segmentation and change detection. In these images, due to the very high spatial resolution, there is a wealth of spatial features within individual objects and among multiple objects, which often span large spatial scales. Additionally, since remote sensing images are captured from a downward-looking camera, the camera can acquire images from any direction, indicating that the spatial features of these images can exist in any direction. Therefore, the ability to globally model the context of VHR remote sensing images and extract large spatial features from multiple directions is essential for dense prediction tasks in VHR remote sensing.

\begin{figure}[!ht]
	\centering
		\includegraphics[width=\linewidth]{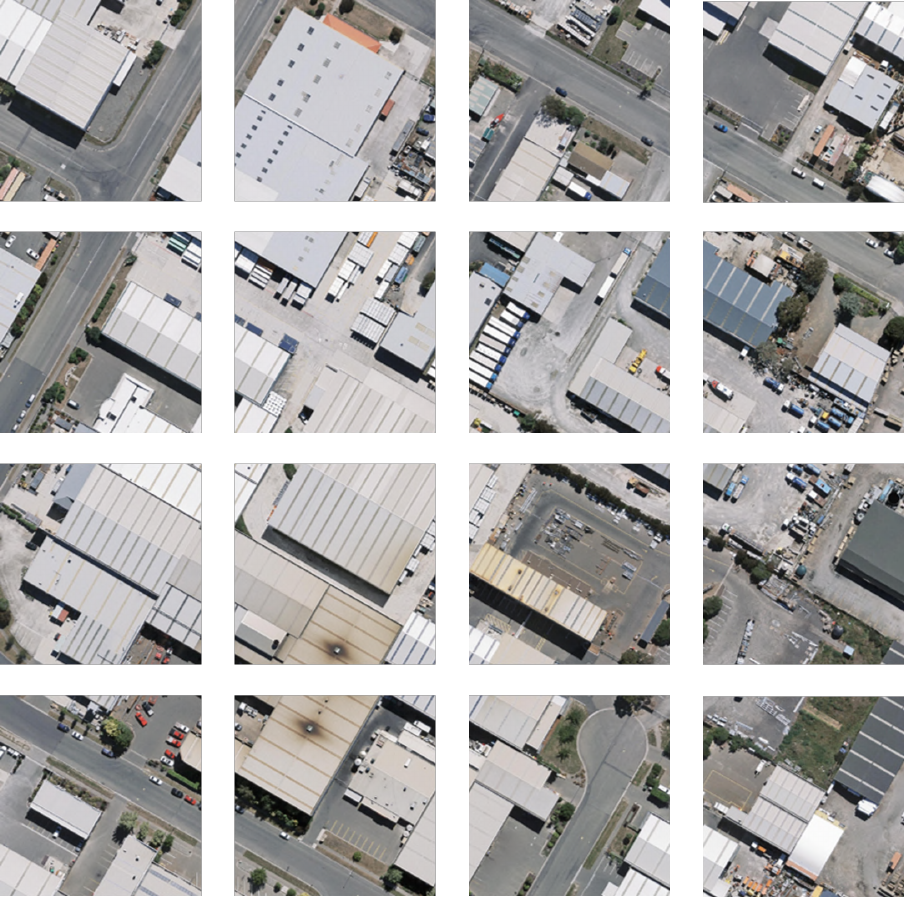}
	\caption{Illustration of the large image preprocessing strategy 
of transformer-based models. Dividing a large VHR remote sensing image into small patches results in the loss of many spatial features. Each patch contains very limited contextual information compared to the original large image.}
 \label{Fig:crop_image}
\end{figure}

In recent years, deep learning models based on transformers~\cite{transformer} have been widely applied to VHR remote sensing dense prediction tasks~\cite{bit, changeformer, roadformer, transformer2}. The transformer architecture, famous for its ability to capture global contextual information and model spatial dependencies effectively through self-attention mechanisms, has achieved impressive results in this domain. However, due to the quadratic complexity of transformers, training and inference with these models on large VHR remote sensing images necessitate dividing these images into smaller patches, as shown in Figure \ref{Fig:crop_image}. This preprocessing step inevitably results in each patch containing only a portion of an object, offering limited contextual information. Consequently, the loss of internal spatial features within individual objects and the spatial dependencies among multiple objects can adversely affect the performance of VHR remote sensing tasks. This limitation underscores the need for innovative solutions that can efficiently process whole images or large segments to preserve and leverage the rich contextual information inherent in large VHR remote sensing images.

The recent work Mamba~\cite{mamba} integrated time-varying parameters into the State Space Model (SSM) and proposed a hardware-aware algorithm that facilitates highly efficient training and inference processes. The SSM, which draws inspiration from the classical Kalman filter model~\cite{kalman}, excels at capturing long-range dependencies and benefits from parallel training capabilities. Research on Mamba has illustrated its potential as a promising alternative to transformers in language modeling due to its robust contextual modeling capacity and linear complexity~\cite{mamba}. However, Mamba is designed to process data along a specific direction, where preceding data cannot establish connections with subsequent data. This directional processing limitation renders it less applicable for image data, which lack a specific orientation and where spatial relationships are crucial across all dimensions.

Recent works such as Vim~\cite{vim} and VMamba~\cite{vmamba} have harnessed SSM to achieve linear complexity and a global effective receptive field, tackling tasks such as image classification and segmentation on natural images. To address the challenge of image data nondirectionality, Vim employs SSM for selective scanning in both the forward and backward directions along the horizontal axis of an image. VMamba extends this approach by conducting selective scanning with SSM in both the horizontal and vertical directions, ensuring that every segment of the image can establish connections with other parts from both the horizontal and vertical directions. The visualization of the effective receptive field in VMamba shows that the effective receptive field is enhanced across both the horizontal and vertical directions~\cite{vmamba}. This indicates that the selective scanning direction of the SSM can significantly impact the effective receptive field in specific directions.

However, Vim and VMamba are not ideally suited for VHR remote sensing images. Objects in natural images are captured from an eye-level view and adhere to the law of gravitation, which means that their main spatial features are distributed in the horizontal and vertical directions. For example, a natural image of a cat sitting on the ground cannot be rotated arbitrarily, as the cat's posture needs to conform to the law of gravity. In contrast, remote sensing images can be rotated freely as they are captured from a top-down satellite perspective, which means that their main spatial features can be distributed in any direction. Given that objects within VHR remote sensing images often span large spatial scales, the spatial features of individual objects and the dependencies among multiple objects can vary in direction. Therefore, VHR remote sensing images contain large spatial features in multiple directions. Due to the impact of the SSM' selective scanning direction on the effective receptive field in specific orientations, Vim's horizontal scanning and VMamba's horizontal and vertical scanning, while effective for natural images with primary features along these axes, cannot adequately address the diverse directional large spatial features inherent in VHR remote sensing images.

To address the aforementioned challenges, we introduce SSM to VHR remote sensing dense prediction tasks for the first time, aiming to achieve global modeling capability and linear complexity. We propose Remote Sensing Mamba (RSM) to process VHR remote sensing images, leveraging the strengths of SSM to extract large and multidirectional spatial features in VHR remote sensing images with rich contextual information.

The RSM can globally model the context of large VHR remote sensing images without self-attention operations. Furthermore, rather than dividing large VHR remote sensing images into small patches, the RSM can handle whole images without losing contextual information due to its linear complexity. Thus, the RSM is suitable for efficiently handling VHR remote sensing images.

Furthermore, we propose the Omnidirectional Selective Scan Module (OSSM) to extract large spatial features from multiple directions in VHR remote sensing images. The OSSM employs SSM for selective scanning in the forward and backward directions across the horizontal, vertical, diagonal, and anti-diagonal axes. This approach enhances the global modeling capability of the contextual information in multiple directions, allowing for the extraction of comprehensive global spatial features. 

In summary, our contributions are as follows:

\begin{enumerate}

 % \item We introduce Remote Sensing Mamba for VHR remote sensing tasks. The RSM first introduced SSM to process VHR remote sensing images, which is capable of handling large VHR remote sensing images with rich contextual information.

 \item We introduce the state space model for the first time to perform dense prediction tasks in VHR remote sensing. The proposed RSM employs SSM to process VHR remote sensing images with linear complexity, enabling direct handling of large remote sensing images without the necessity of segmenting them into small patches, which preserves the rich contextual information inherent in remote sensing images. % Moreover, RSM leverages SSMs for global modeling of VHR remote sensing images, utilizing the global contextual information in large remote sensing images to perform dense prediction tasks.

 % \item We design an Omnidirectional Selective Scan Module to extract large spatial features in various directions. As the spatial features in VHR remote senisng images span large spatial scales and multiple directions, OSSM selective scan images across multiple directions, thus extracting global spatial features in multiple directions.

 % \item We design an Omnidirectional Selective Scan Module (OSSM) to extract large spatial features in multiple directions. Considering the large spatial scales and multiple directions of spatial features in VHR remote sensing images, OSSM utilizes SSM to selectively scan remote sensing images in multiple directions, thereby globally modeling remote sensing images and extracting large spatial features from multiple directions.

 \item We design an Omnidirectional Selective Scan Module (OSSM) to extract large spatial features in multiple directions. Considering that the land covers can be distributed in any spatial direction as remote sensing images are captured from a top-down perspective, RSM utilizes OSSM to selectively scan remote sensing images in multiple directions, thereby modeling the global context of remote sensing images.

 \item We demonstrate the efficiency and superiority of the RSM in VHR remote sensing tasks. Experiments on the semantic segmentation datasets (WHU and Massachusetts Road) and the change detection datasets (LEVIR-CD and WHU-CD) show that the RSM achieves state-of-the-art performance on both semantic segmentation and change detection tasks.
\end{enumerate}

\section{Related Works}

\subsection{Dense Prediction of Very High Resolution Remote Sensing}

Dense prediction tasks in VHR remote sensing mainly include semantic segmentation and change detection tasks, where remote sensing images have very high spatial resolution and abundant contextual information. Large remote sensing images span large spatial ranges, thereby offering rich contextual information, which is crucial for dense prediction tasks. Experiments in FCCDN~\cite{fccdn} and SwinB-CNN~\cite{hybrid1} show that models utilizing larger remote sensing images can achieve superior performance in dense prediction tasks. This underscores the importance of leveraging rich contextual information in large remote sensing images for dense prediction tasks.

There are three main types of deep learning models for VHR remote sensing dense prediction tasks: Convolutional Neural Network (CNN) based models, transformer-based models and CNN-transformer hybrid based models. 

CNN-based models excel in image processing due to their ability to efficiently capture local spatial features through their hierarchical structure. Papadomanolaki et al.~\cite{cnn1} proposed an urban change detection framework that combines U-Net~\cite{unet} for feature extraction and LSTMs~\cite{lstm} for temporal modeling. Yue et al.~\cite{treeunet} proposed an adaptive network to increase the classification rate at the pixel level based on a deep semantic model infrastructure. Zhao et al.~\cite{cnn2} introduced a novel change detection framework named EDED, which operates by exchanging features between two encoder branches. This approach enables the separate identification of changed objects in bitemporal images to produce change detection results. CNN-based models often incorporate attention modules to focus on important areas within images, thereby enhancing feature extraction capabilities. Yang et al.~\cite{attention_cnn1} proposed a multipath attention-fused
block module to fuse multipath features, and a refinement attention-fused block module to fuse high-level abstract features and low-level spatial features. Han et al.~\cite{attention_cnn2} proposed a change guide module for change detection, which can effectively capture the long-distance dependency among pixels and effectively overcomes the problem of the insufficient receptive field of traditional convolutional neural networks. To address the issue of varying spatial scales of objects in remote sensing images, CNN-based models utilize multi-scale modules to extract features at multiple scales, thus extracting object features of varying spatial scales. Diakogiannis et al.~\cite{resunet} incorporated atrous convolutions and pyramid scene parsing pooling into a UNet backbone, extracting multi-scale features of remote sensing images. Han et al.~\cite{multiscale2} used hierarchical convolution operations to extract multi-scale features, continuously merges multi-scale features layer by layer to improve the expression of global and local information. Gu et al.~\cite{cnn3} focused on exploiting the multiscale feature differences between bitemporal images to concentrate on the detailed information of the changing areas.

However, the inherently large spatial scales of objects in VHR remote sensing images pose a challenge for CNN models. Due to their limited ability to capture global receptive fields, CNNs struggle to extract comprehensive global spatial features and dependencies within these images. Conversely, transformers excel at global context modeling across entire images by using self-attention mechanisms, thereby overcoming the limitations of CNNs in capturing global spatial relationships. This attribute has led to the widespread application of transformer-based models in VHR remote sensing tasks, demonstrating their ability to perform semantic segmentation and change detection with an enhanced understanding of the global context. Bandara et al.~\cite{changeformer} proposed a transformer-based Siamese network architecture for change detection, which unified a hierarchically structured transformer encoder with a multi-layer perception decoder in a Siamese network architecture to efficiently render multi-scale long-range details. Zhang et al.~\cite{transformer2} proposed a purely transformer-based architecture for change detection tasks, constructing a model based on the swin transformer~\cite{swintransformer} architecture. Transformer-based models also model both local and global contexts, leveraging local and global features for dense prediction tasks. Liu et al.~\cite{transformer_lg} adopted a global-local attention module containing global attention blocks and local attention blocks to capture the global and local context information and fully exploit the global and local features. Jiang et al.~\cite{roadformer} designed a multi-context patch embedding scheme, which adopted a multi-range, multi-view context observation strategy to obtain higher-quality token embedding.

The global modeling capabilities of transformer-based models have led to their widespread application in dense prediction tasks for remote sensing. However, their quadratic complexity and substantial computational demands pose challenges for model training and inference. In addition to the global features of remote sensing images, local features are equally important for dense prediction tasks. Consequently, CNN-transformer hybrid-based models have been introduced and extensively applied to dense prediction tasks in remote sensing. On the one hand, hybrid-based models can reduce the number of parameters and computations compared to transformer-based models, which facilitates more efficient training and inference. Zhang et al.~\cite{hybrid1} proposed an encoder–decoder network for remote sensing image segmentation, in which the encoder module uses a swin transformer~\cite{swintransformer} to extract features to achieve better long-range spatial dependencies modeling, and the decoder module draws on some effective blocks and successful strategies of CNN-based models. Chen et al.~\cite{bit} integrated transformers into change detection tasks, utilizing a ResNet~\cite{resnet} as the encoder and self-attention modules as decoders. On the other hand, these models are capable of leveraging both the local features extracted by CNNs and the global features extracted by transformers, thus paying attention to both local and global aspects of images for dense prediction tasks. Li et al.~\cite{transunetcd} proposed an encoding–decoding hybrid transformer model for change detection, which has the advantages of both transformers and UNet~\cite{unet} in learning global context and low-level details. Li et al.~\cite{hybrid2} proposed a CNN–transformer network with multiscale framework to better exploit global–local information in optical remote sensing images.

Despite their strengths, transformer-based models and hybrid-based models still encounter challenges due to the quadratic complexity of their self-attention mechanisms, particularly when processing large VHR remote sensing images. This necessitates transformer-based models and hybrid-based models to divide large images into smaller segments, which can result in a significant loss of contextual information. While transformers can model global contexts, the reduced contextual information limits their effectiveness in VHR remote sensing dense precition tasks. In response, we propose the SSM-based RSM for VHR remote sensing dense prediction tasks, which has linear complexity and global modeling capability. The RSM is adept at handling large images with rich contextual information, thereby providing a more effective solution for processing VHR remote sensing images.

\subsection{State Space Models}

State Space Models (SSM) have gained significant traction in the field of deep learning in recent years, marking a remarkable evolution in the way long-range dependencies and sequential data are handled~\cite{s4, lssl, ssmgood3}. Initially inspired by their traditional application in control systems, SSMs were innovatively adapted to deep learning, leveraging the strengths of continuous state spaces to model complex temporal dynamics. The integration of SSM into deep learning was catalyzed by the introduction of the Highest Polynomial Powered Operator (HiPPO) initialization method~\cite{hippo}, which significantly improved the models' ability to capture long-range dependencies.

The LSSL model demonstrated the potential of SSM in addressing long-range dependency challenges, setting a foundation for subsequent research in the field~\cite{lssl}. However, LSSL faces critical hurdles related to computational and memory efficiency, limiting its practical application. Addressing these limitations, the S4 model introduced by Gu et al. emerged as a pivotal advancement, proposing a normalized parameterization strategy that reduced computational overhead, thereby making SSM more feasible for practical applications~\cite{s4}.

Following the breakthrough of the S4 model, the landscape of SSM research expanded rapidly, with several variants being developed to enhance the model's structure and efficiency. Notable among these are models incorporating complex-diagonal structures to improve temporal modeling capabilities~\cite{temporal_modeling1, temporal_modeling2}, as well as those supporting multiple-input multiple-output configurations to increase model flexibility~\cite{ssmgood3}. Additionally, innovations such as the decomposition of operations into diagonal plus low-rank structures~\cite{low_rank} and the introduction of selection mechanisms~\cite{mamba} have further refined the adaptability and performance of SSMs.

However, the aforementioned models are only capable of processing unidirectional sequence data and cannot handle image data that lack a specific direction. Recent works~\cite{vim, vmamba, changemamba} have achieved global modeling of the context of images using SSM by conducting selective scanning both forward and backward in certain directions. Vim~\cite{vim} performs selective scanning in the horizontal direction, enabling each part of the image to perceive global information. VMamba~\cite{vmamba} extends this by conducting selective scanning both horizontally and vertically, enhancing the model's global effective receptive field in both dimensions.

Nevertheless, since the primary spatial features of natural images are distributed in both the horizontal and vertical directions, and VHR remote sensing images exhibit large spatial features in multiple directions, although Vim and VMamba achieved good performance on natural images, they are not suitable for VHR remote sensing images. The proposed Omnidirectional Selective Scan Module conducts selective scanning in multiple directions, and is capable of capturing the large spatial features of VHR remote sensing images in various directions.

\section{Methodology}

\subsection{Preliminaries: State Space Models}
\label{section:3.1}

In the realm of deep learning, State Space Models (SSMs) have gained prominence for their ability to encapsulate dynamic systems that map an input sequence \( x(t) \in \mathbb{R}^L \) to an output \( y(t) \in \mathbb{R} \). SSMs are grounded in the principles of control theory and are defined by a set of linear ordinary differential equations (ODEs):

\begin{equation}
h'(t) = A h(t) + B x(t)
\end{equation}

\begin{equation}
y(t) = C h(t) + D x(t)
\end{equation}

where \( A \in \mathbb{C}^{N \times N} \), \( B \in \mathbb{R}^{N \times L} \), \( C \in \mathbb{R}^{N} \), and \( D \in \mathbb{R}^{L} \) are the system matrices and \( h(t) \) denotes the hidden state vector at time \( t \).

The model's state-transition matrix \( A \) governs the evolution of the state vector \( h(t) \), while the input matrix \( B \), output matrix \( C \), and feedthrough matrix \( D \) articulate the relationships between the input \( x(t) \), state \( h(t) \), and output \( y(t) \), respectively. In discrete-time settings, which are typical in deep learning applications, these continuous equations must be discretized for computational tractability and alignment with data sampling rates.

The discretization of SSM involves transforming the continuous ODE into a discrete-time representation. Employing a zero-order hold on the input signal, the discrete-time SSM can be represented as:

\begin{equation}
h_k = \Phi h_{k-1} + \Gamma x_k
\end{equation}

\begin{equation}
y_k = C h_k + D x_k
\end{equation}

where \( h_k \) is the hidden state at discrete time step \( k \), \( y_k \) is the output, \( \Phi = e^{A \Delta T} \) is the state transition matrix for time step \( \Delta T \), and \( \Gamma \) is derived as \( \Gamma = (e^{A \Delta T} - I) A^{-1} B \), assuming that the input remains constant over each interval \( \Delta T \).

The Mamba~\cite{mamba} methodology distinguishes itself within the SSM framework by adopting a selective scan mechanism. This mechanism enhances the standard SSM structure by permitting dynamic adjustments to system matrices \( B \) and \( D \), based on the current and historical context of the input sequence. Consequently, Mamba's SSM can model complex temporal dynamics more effectively, as these matrices adapt in response to the evolving features of the input data.

\subsection{Overall Architecture}
\label{section:3.2}

\begin{figure*}[!ht]
 \centering
 	\includegraphics[width=\linewidth]{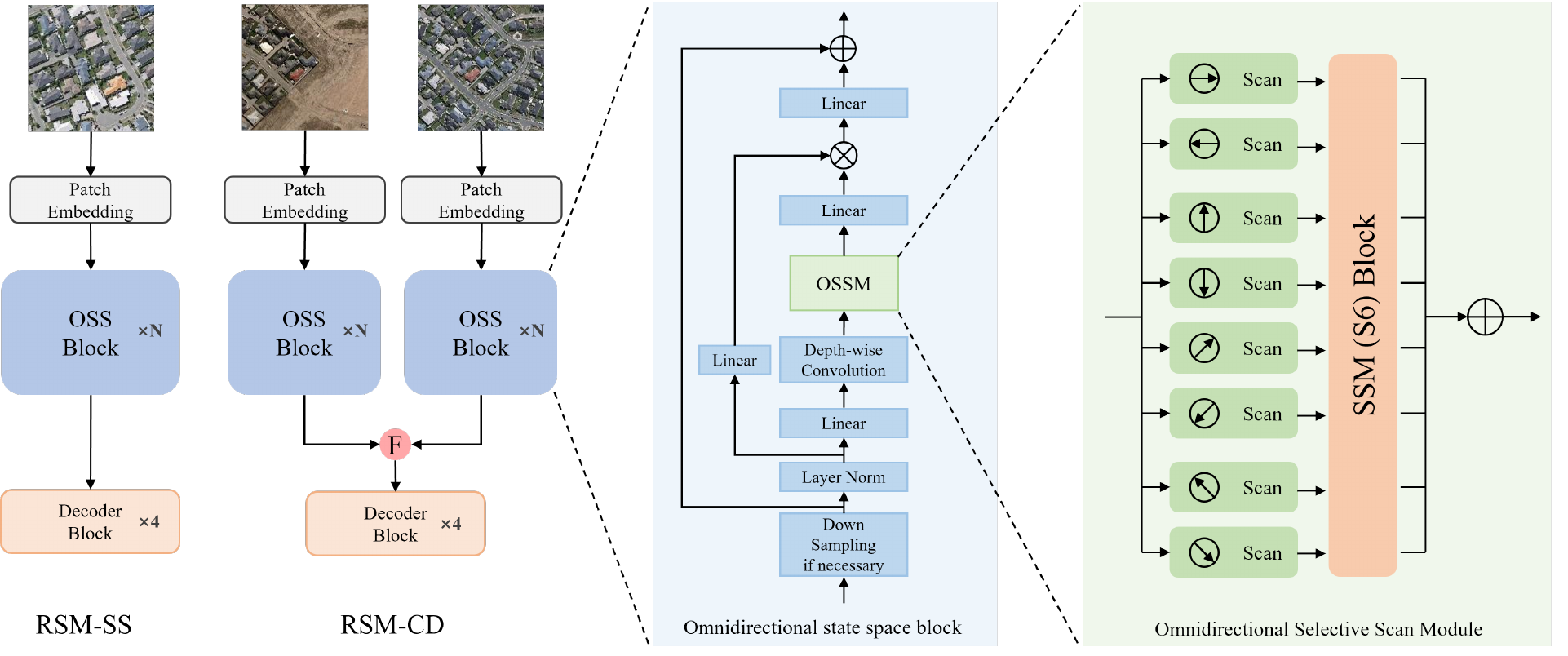}
 \caption{Illustration of the Overall structure of RSM-SS and RSM-CD. RSM-SS and RSM-CD can globally model the context of images in multiple directions with linear complexity using the omnidirectional selective scan.}
 \label{Fig:rsm_structure}
\end{figure*}

VHR remote sensing images are primarily utilized in semantic segmentation and change detection tasks. Consequently, we have developed two specialized frameworks: Remote Sensing Mamba for Semantic Segmentation (RSM-SS) for the semantic segmentation task and Remote Sensing Mamba for Change Detection (RSM-CD) for the change detection task, as illustrated in Figure \ref{Fig:rsm_structure}. To demonstrate the effectiveness of SSM in processing VHR remote sensing images, RSM-SS and RSM-CD employ the simplest architectures for semantic segmentation and change detection, respectively. Our objective is to show that RSM can achieve state-of-the-art performance with the simplest architectures, thus demonstrating the substantial potential of SSM-based models for remote sensing dense prediction tasks.

%%%
The RSM-SS architecture utilizes the U-Net~\cite{unet} encoder-decoder framework, where input VHR remote sensing images are first transformed into a sequence of image patches through patch embedding. These patches are then fed into the encoder to extract features, which are subsequently upsampled by the encoder to produce semantic segmentation results. The encoder consists of five stages, each comprising several OSS blocks. Stage 1 extracts features from the input VHR remote sensing images, while stages 2-5 progressively downsample the encoder features and double the number of feature channels at each stage. The encoder is composed of four encoder blocks, where features are upsampled and then concatenated with the encoder features along the channel dimension through skip connections followed by convolution. This process fuses the semantic information of decoder features with the spatial information of encoder features, facilitating semantic segmentation from both global and local perspectives.

The RSM-CD employs an FC-Siam-Conc~\cite{fcef} Siamese network architecture. Bitemporal VHR remote sensing images are first converted into bitemporal sequences of image patches using patch embedding, which are then fed into bitemporal encoders with shared weights to extract features. These bitemporal encoder features are simply fused and upsampled in a single encoder to obtain the final change detection results. Similar to RSM-SS, the shared-weight encoders in RSM-CD consist of five stages with several OSS blocks each, and the encoder comprises four encoder blocks. After feature extraction by the shared-weight encoders, bitemporal features of the same size are concatenated along the channel dimension and convolved. This fusion captures the information of both temporal phases of VHR remote sensing images, enabling the effective segmentation of changed objects. The fused features are upsampled in the encoder and concatenated with fusion features of the same size through skip connections and convolution, thus preserving rich semantic and spatial information.

\subsection{Omnidirectional State Space Block}
\label{section:3.3}

The Omnidirectional State Space (OSS) block is a novel feature extraction unit designed for semantic segmentation and change detection tasks in VHR remote sensing images, as shown in Figure \ref{Fig:rsm_structure}. Central to the OSS block is the Oriented Scanning Module (OSSM), which serves as the core for global contextual modeling across multiple orientations within an image. The OSSM selectively scans the input image in various directions, capturing the intricate spatial relationships and providing a comprehensive understanding of the context.

The architecture begins with a layer normalization that standardizes the input data, enhancing the model training stability. Following this, a linear transformation adjusts the dimensionality of the data, preparing it for the depth-wise convolution process. This convolution operates on each input channel separately, reducing the parameter count and focusing on extracting spatial features. After convolution, the features pass through the OSSM. The OSSM performs selective scanning on the features in the forward and backward directions along the horizontal, vertical, diagonal, and anti-diagonal directions, which are then added together. The output from the OSSM then undergoes a linear transformation and a gating operation, adjusting deep features with outputs of a linear transformation of the normalized features. Finally, the features are passed to a final linear layer, which are then added with input features through a residual connection.

The OSS block is crafted with a keen focus on balancing computational efficiency and the ability to extract rich spatial features from VHR remote sensing images. Therefore, as the OSS block is efficient and lightweight, we can stack more blocks with a similar total model depth budget when building the model.

\subsection{Omnidirectional Selective Scan Module}
\label{section:3.4}

\begin{figure}[!ht]
	\centering
		\includegraphics[width=\linewidth]{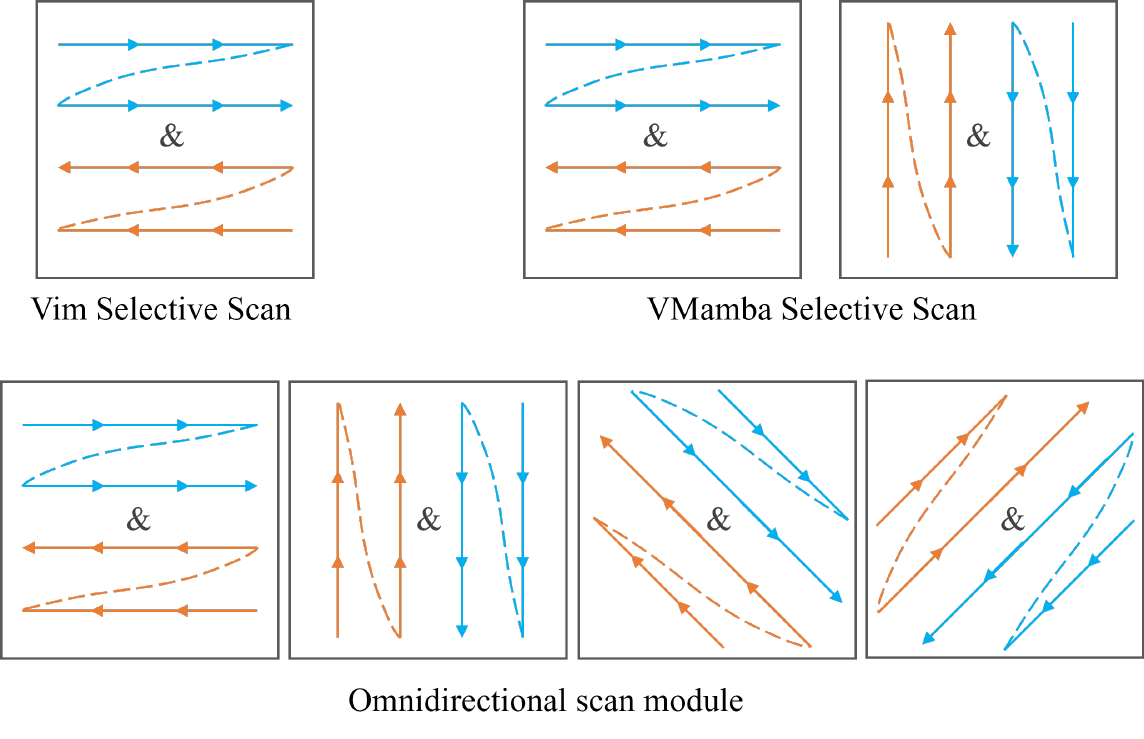}
	\caption{Illustration of the selective scan directions of Vim, VMamba, and OSSM.}
 \label{Fig:ossm_comparsion}
\end{figure}

Vim and VMamba have demonstrated commendable performance in natural images, where primary spatial features are distributed along the horizontal and vertical directions. Vim conducts selective scanning in the forward and backward directions along the horizontal axis, and VMamba introduces selective scanning across both the horizontal and vertical directions, allowing every part of the image to globally attend in both forward and backward directions along a specific direction, as illustrated in Figure \ref{Fig:ossm_comparsion}.

Specifically, the structure of the OSSM is illustrated in Figure 5. OSSM begins with the input image patches undergoing omnidirectional scanning in the horizontal, vertical, diagonal, anti-diagonal directions, and their respective reverse directions, resulting in eight sequences of image patches. These sequences are then stacked along a new dimension and fed into the S6 block. The S6 block's selective scanning mechanism independently processes each image patch sequence, performing global modeling in specific directions~\cite{mamba}. Finally, all the image patch sequences are added after unstacking, merging global modeling information from multiple directions. This method enables the extraction of large spatial features from various orientations within VHR remote sensing images.

\begin{figure*}[!ht]
	\centering
		\includegraphics[width=\linewidth]{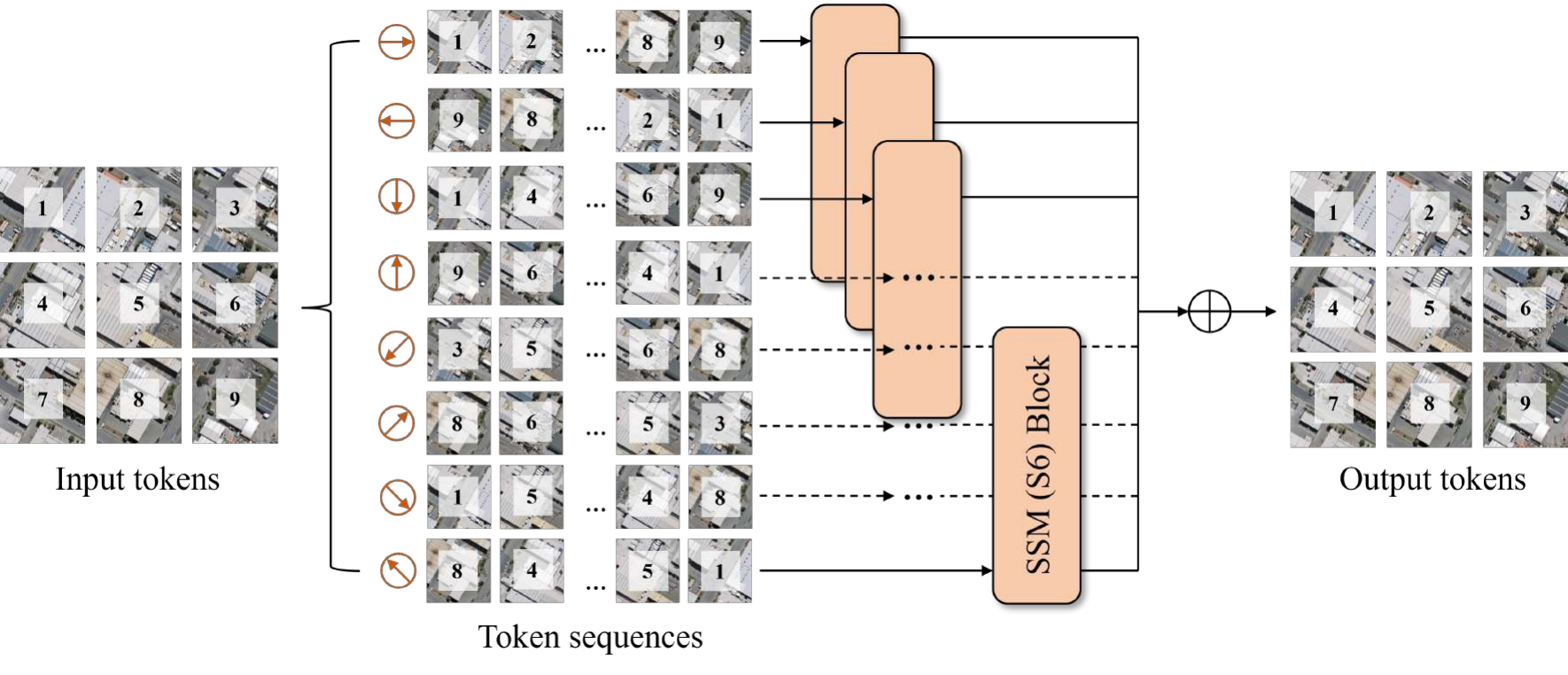}
	\caption{Illustration of the structure of the omnidirectional selective scan module.}
 \label{Fig:ossm_structure}
\end{figure*}

% ---------------------------------------

\section{Experimental Settings and Results}

To validate the efficiency and superiority of the RSM in VHR remote sensing tasks, we conducted experiments across two distinct tasks: semantic segmentation and change detection. For the semantic segmentation task, we evaluated the effectiveness of the RSM-SS model on the WHU~\cite{whu} dataset and the Massachusetts Road~\cite{mszs} dataset. For the change detection task, we evaluated the effectiveness of the RSM-CD model on the WHU-CD~\cite{whu} dataset and the LEVIR-CD~\cite{levircd} dataset.

\subsection{Datasets}

We offer a brief description of the experimental 
semantic segmentation and change detection datasets in Table \ref{datasets_introduction}. 

\begin{table}[!ht]
\caption{Brief Introduction of The Experimental Datasets.}
\label{datasets_introduction}
\centering
\begin{tabular}{lccccc}
\toprule
Name & Task & Resolution (m) & Images & Image size \\
\midrule
WHU~\cite{whu} & Seg & 0.3 & 8189 & 512×512 \\
M-Road~\cite{mszs} & Seg & 1 & 1171 & 1500 × 1500 \\
WHU-CD~\cite{whu} & CD & 0.075 & 1 & 32207×15354 \\
LEVIR-CD~\cite{levircd} & CD & 0.5 & 637 & 1024×1024 \\
\bottomrule
\end{tabular}
\end{table}

\subsubsection{Semantic Segmentation Datasets}

The WHU~\cite{whu} building dataset is composed of two distinct subsets: one featuring satellite images and the other showcasing aerial photographs. Our investigation employs the aerial images subset, which includes a total of 8,189 images. The images are divided into 4,736 images designated for training, 1,036 for validation, and 2,416 for testing purposes, each with a spatial resolution of 0.3 meters. This subset collectively captures approximately 22,000 buildings across an expanse of more than 450 square kilometers.

The Massachusetts~\cite{mszs} Roads dataset incorporates 1,171 aerial photographs from Massachusetts, with each image measuring 1500×1500 pixels and encompassing 2.25 square kilometers. This dataset is organized into 1,108 training images, 14 validation images, and 49 test images. It encompasses a diverse array of environments, including urban, suburban, and rural areas, spanning more than 2,600 square kilometers, with the test segment covering 
more than 110 square kilometers. For analytical purposes, we segment the images into 1024 × 1024-pixel patches with a 548-pixel overlap on both the horizontal and vertical axes.

\subsubsection{Change Detection Datasets}

The WHU-CD~\cite{whu} dataset includes bitemporal VHR aerial images from 2012 and 2016, revealing significant alterations in building structures. We segment the dataset into 1024 × 1024 pixel patches that do not overlap and distribute these patches into training, validation, and test sets at a 7:1:2 ratio.

The LEVIR-CD~\cite{levircd} dataset is an extensive change detection dataset comprising VHR (0.5 m/pixel) Google Earth images that document a range of building transformations over a period of 5 to 14 years. This dataset is particularly focused on changes related to buildings, such as construction and demolition. The bitemporal images are meticulously labeled with binary masks by specialists, indicating changes (1) and no changes (0), featuring a total of 31,333 instances of building changes. We segment the dataset into nonoverlapping 1024× 1024 pixel patches.

\subsection{Benchmark Methods}

To evaluate the effectiveness of the proposed Remote Sensing Mamba, we conducted comparative experiments with various benchmark methods on the semantic segmentation and change detection tasks. The benchmark methods tested on the same dataset are based on the same splitting of the dataset and use the same data.

On the semantic segmentation task, the compared CNN-based models include FCN~\cite{fcn}, SegNet~\cite{segnet}, U-Net~\cite{unet}, PSPNet~\cite{psp}, HRNet~\cite{hrnet}, MA-FCN~\cite{mafcn}, Deeplabv3+~\cite{deeplab}, ResUNet~\cite{resunet_model}, MAP-Net~\cite{mapnet}, D-LinkNet~\cite{dlinknet} and SIINet~\cite{siinet}, the compared transformer-based models include Segformer~\cite{segformer} and RoadFormer~\cite{roadformer} , and the CNN-transformer hybrid models include BDTNet~\cite{bdtnet}, TransUNet~\cite{transunet} and CMTFNet~\cite{cmtfnet}.

On the change detection task, the compared CNN-based models include FC-EF~\cite{fcef}, FC-Siam-Diff~\cite{fcef}, FC-Siam-Conc~\cite{fcef}, STANet~\cite{sta}, DTCDSCN~\cite{dtcdstn}, SNUNet~\cite{snu}, CDNet~\cite{cdnet}, DDCNN~\cite{ddcnn}, DASNet~\cite{dasnet}, DSIFN~\cite{dsifn} and HANet~\cite{hanet}, the compared transformer-based models include ChangeFormer~\cite{changeformer}, and the CNN-transformer hybrid models include BIT~\cite{bit}, MTCNet~\cite{mtcnet}, MSCANet~\cite{mscanet} and AMTNet-50~\cite{amtnet}.

\subsection{Implementation Details}

\subsubsection{Data Augmentation}

To demonstrate the effectiveness of the proposed methods, we only employed straightforward data augmentation techniques, avoiding the use of any elaborate tricks. For the semantic segmentation task, the data augmentation methods used for the RSM-SS model included flipping (p=0.5) and transposing (p=0.5). For the change detection task, the data augmentation methods used for the RSM-CD model included flipping (p=0.5), transposing (p=0.5), and swapping of bitemporal images (p=0.5). 

\subsubsection{Training and Inference}
\label{section:4.3.2}

We utilized PyTorch~\cite{pytorch} to construct and deploy RSM-SS and RSM-CD on a single RTX A100 GPU. Given the variable image sizes across datasets, we adjusted the batch sizes accordingly: 16 for the WHU dataset, 4 for both the Massachusetts Roads and WHU-CD datasets, and 64 for the LEVIR-CD dataset. Our loss function integrates binary cross-entropy loss with Dice coefficient loss to optimize performance. We employed the AdamW~\cite{adam} optimizer with an initial learning rate of 0.001 and a weight decay of 0.001. The learning rate adjustment strategy is to reduce the learning rate by a factor of 0.1 if there is no improvement in the F1-score on the validation set over a span of 10 epochs. The models were trained over 150 epochs to ensure ample training and convergence. Checkpoints capturing the highest F1-scores on the validation sets were preserved for subsequent testing. To maintain consistency with other change detection methodologies, we initialized our models using the default settings provided by PyTorch for all datasets.

\subsubsection{Evaluation Metrics}

To evaluate the performance of the proposed models, we employ four key evaluation metrics: precision (P), recall (R), F1-score, and intersection over union (IoU). Precision quantifies the rate of false positives within the results, whereas recall measures the rate of false negatives. Achieving high scores in both precision and recall simultaneously poses a significant challenge due to their inversely proportional relationship. The F1-score, which represents the harmonic mean of precision and recall, serves as a balance between the two by simultaneously considering both metrics. Additionally, the IoU metric measures the proportion of overlap between the predicted and actual changed pixels relative to the total area of union, providing a spatial accuracy assessment of the model’s predictions.

\subsection{Ablation Study}
\label{section:4.4.1}

To verify the effectiveness of OSSM, comparative experiments were conducted on the Massachusetts Roads dataset for the semantic segmentation task and the WHU-CD dataset for the change detection task. We compared the performance of three variations: SS1D~\cite{vim}, which employs selective scanning in the horizontal direction and its reverse direction; SS2D~\cite{vmamba}, which includes selective scanning in both the horizontal and vertical directions and their reverses; and OSSM, which extends selective scanning to eight directions—horizontal, vertical, diagonal, and anti-diagonal, along with their reverse directions. This comparison aimed to demonstrate the superiority of employing eight-directional selective scanning in VHR remote sensing images.

Table \ref{ossm_ablation} presents the comparative results of SS1D, SS2D, and OSSM, indicating that OSSM outperforms both SS1D and SS2D in semantic segmentation and change detection tasks. Specifically, for the semantic segmentation task on the Massachusetts Roads dataset, the presence of roads extending in multiple directions with significant spatial scales necessitates selective scanning across multiple directions, thereby extracting large road features oriented in various directions. Similarly, in the change detection task on the WHU-CD dataset, the spatial features of buildings, such as edge characteristics and arrangement directions, require selective scanning in multiple directions to capture large architectural features from various orientations. Compared to SS1D and SS2D, the omnidirectional selective scan of OSSM enables the extraction of large object features from multiple directions, making it more suitable for VHR remote sensing images.

\begin{table}[!ht]
\caption{Ablation Study of OSSM on the Massachusetts Road dataset and WHU-CD dataset. The Best Values Are Highlighted in Bold in Each Dataset.}
\label{ossm_ablation}
\centering
\begin{tabular}{lcccccc} 
\toprule
Dataset & Task & Module & P (\%) & R (\%) & F1 (\%) & IoU (\%) \\
\midrule
M-Road & Seg & SS1D & 85.17 & 74.37 & 79.40 & 65.84 \\
M-Road & Seg & SS2D & 85.57 & 74.78 & 79.81 & 66.41 \\
M-Road & Seg & OSSM & \textbf{86.52} & \textbf{75.24} & \textbf{80.49} & \textbf{67.35} \\
\midrule
WHU-CD & CD & SS1D & 91.86 & 89.33 & 90.58 & 82.78 \\
WHU-CD & CD & SS2D & 92.25 & 89.66 & 90.94 & 83.38 \\
WHU-CD & CD & OSSM & \textbf{93.37} & \textbf{90.42} & \textbf{91.87} & \textbf{84.96} \\
\bottomrule
\end{tabular}
\end{table}

% \subsection{Experimental Results}
\subsection{Overall Comparison}

\subsubsection{Semantic Segmentation Task}
\label{section:4.4.3}

This subsection presents the results of comparing RSM-SS with other models on the semantic segmentation task on two datasets (the Massachusetts Road and WHU datasets).

The accuracy comparison results on the Massachusetts Road dataset are presented in Table \ref{Massachusetts_result}. The RSM-SS surpasses all the compared models and achieves the highest IoU (0.6735) and F1-score (0.8049) on the Massachusetts Road dataset. The Massachusetts Road dataset is characterized by roads with extensive spatial scales, where contextual information plays a critical role in road semantic segmentation. Due to its linear complexity, RSM-SS is adept at processing VHR remote sensing images with rich contextual information. This capability allows it to accurately segment roads by effectively leveraging the vast amount of contextual information in remote sensing images.

\begin{table}[!ht]
\caption{Accuracy Comparison on the Massachusetts Road Dataset. The Best Values Are Highlighted in Bold.}
\label{Massachusetts_result}
\centering
\begin{tabular}{lcccc}
\toprule
Methods & P (\%) & R (\%) & F1 (\%) & IoU (\%) \\
\midrule
SegNet~\cite{segnet} & 76.09 & 78.23 & 77.15 & 62.79 \\
U-Net~\cite{unet} & 77.53 & 77.82 & 77.67 & 63.50 \\
ResUNet~\cite{resunet} & 78.77 & 77.45 & 78.10 & 64.07 \\ 
D-LinkNet~\cite{dlinknet} & 78.34 & 77.91 & 78.12 & 64.10 \\
HRNetv2~\cite{hrnet} & 79.01 & 78.20 & 78.60 & 64.75 \\
Deeplabv3+~\cite{deeplab} & 75.14 & 72.56 & 73.83 & 58.51 \\ 
SIINet~\cite{siinet} & 85.36 & 74.13 & 79.35 & 65.77 \\
RoadFormer~\cite{roadformer} & 80.54 & \textbf{78.90} & 79.71 & 66.27 \\
BDTNet~\cite{bdtnet} & 82.99 & 76.37 & 79.54 & 66.03 \\
\midrule
RSM-SS & \textbf{86.52} & 75.24 & \textbf{80.49} & \textbf{67.35} \\
\bottomrule
\end{tabular}
\end{table}

The accuracy comparison results on the WHU dataset are summarized in Table \ref{whu_result}. The accuracy comparison results show that the RSM-SS achieves the highest IoU (0.9081) and F1-score (0.9518) on this dataset. In the WHU dataset, buildings are arranged in a variety of orientations and cover large spatial scales. Extracting large spatial features of buildings in multiple directions is crucial for accurate building detection. RSM-SS performs selective scanning across multiple directions on VHR remote sensing images, which enables the extraction of substantial spatial features of buildings from various angles, thus achieving precise segmentation of buildings in VHR remote sensing images.

\begin{table}[!ht]
\caption{Accuracy Comparison on the WHU Dataset. The Best Values Are Highlighted in Bold.}
\label{whu_result}
\centering
% \scalebox{0.8}{
\begin{tabular}{lcccc}
\toprule
Methods & P (\%) & R (\%) & F1 (\%) & IoU (\%) \\
\midrule
FCN~\cite{fcn} & 92.29 & 92.84 & 92.56 & 86.16 \\
SegNet~\cite{segnet} & 93.42 & 91.71 & 92.56 & 86.15 \\
U-Net~\cite{unet} & 94.50 & 90.88 & 92.65 & 86.31 \\
PSPNet~\cite{psp} & 93.19 & 94.21 & 93.70 & 88.14 \\
HRNet~\cite{hrnet} & 91.69 & 92.85 & 92.27 & 85.64 \\
MA-FCN~\cite{mafcn} & 94.75 & 94.92 & 94.83 & 90.18 \\
Deeplabv3+~\cite{deeplab} & 94.31 & 94.53 & 94.42 & 89.43 \\ 
ResUNet~\cite{resunet} & 94.49 & 94.71 & 94.60 & 89.75 \\
MAP-Net~\cite{mapnet} & 93.99 & 94.82 & 94.40 & 89.40 \\
Segformer~\cite{segformer} & 94.72 & 94.42 & 94.57 & 89.70 \\ 
TransUNet~\cite{transunet} & 94.05 & 93.07 & 93.56 & 87.89 \\
CMTFNet~\cite{cmtfnet} & 90.12 & \textbf{95.21} & 92.59 & 86.21 \\
\midrule
RSM-SS & \textbf{95.25} & 95.12 & \textbf{95.18} & \textbf{90.81} \\
\bottomrule
\end{tabular}
% }
\end{table}

\begin{figure}[!ht] %子图加并列
\centering
\subfigure[]{
\begin{minipage}[b]{0.25\linewidth}
\includegraphics[width=\linewidth]{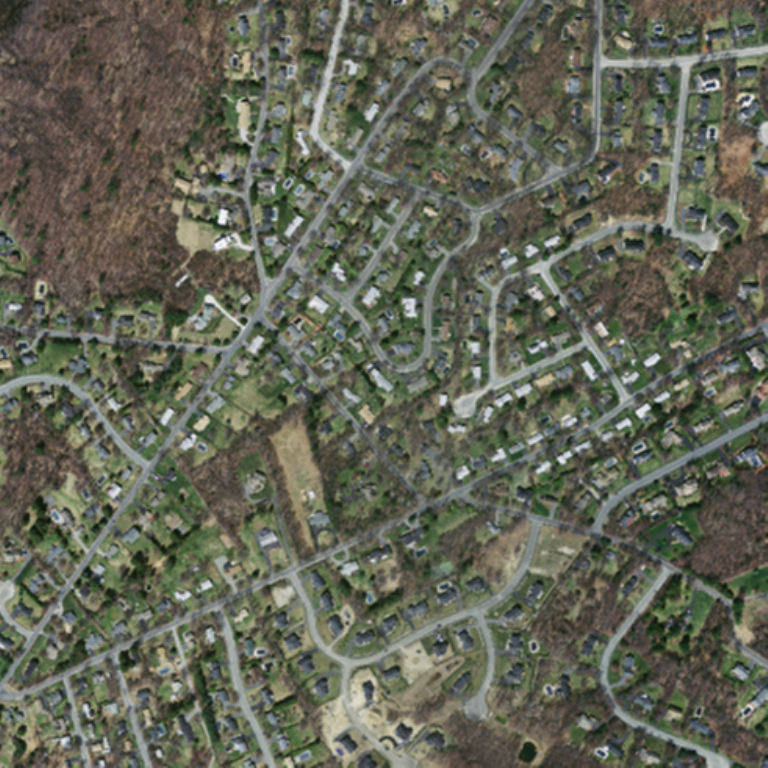} \\ \\
\includegraphics[width=\linewidth]{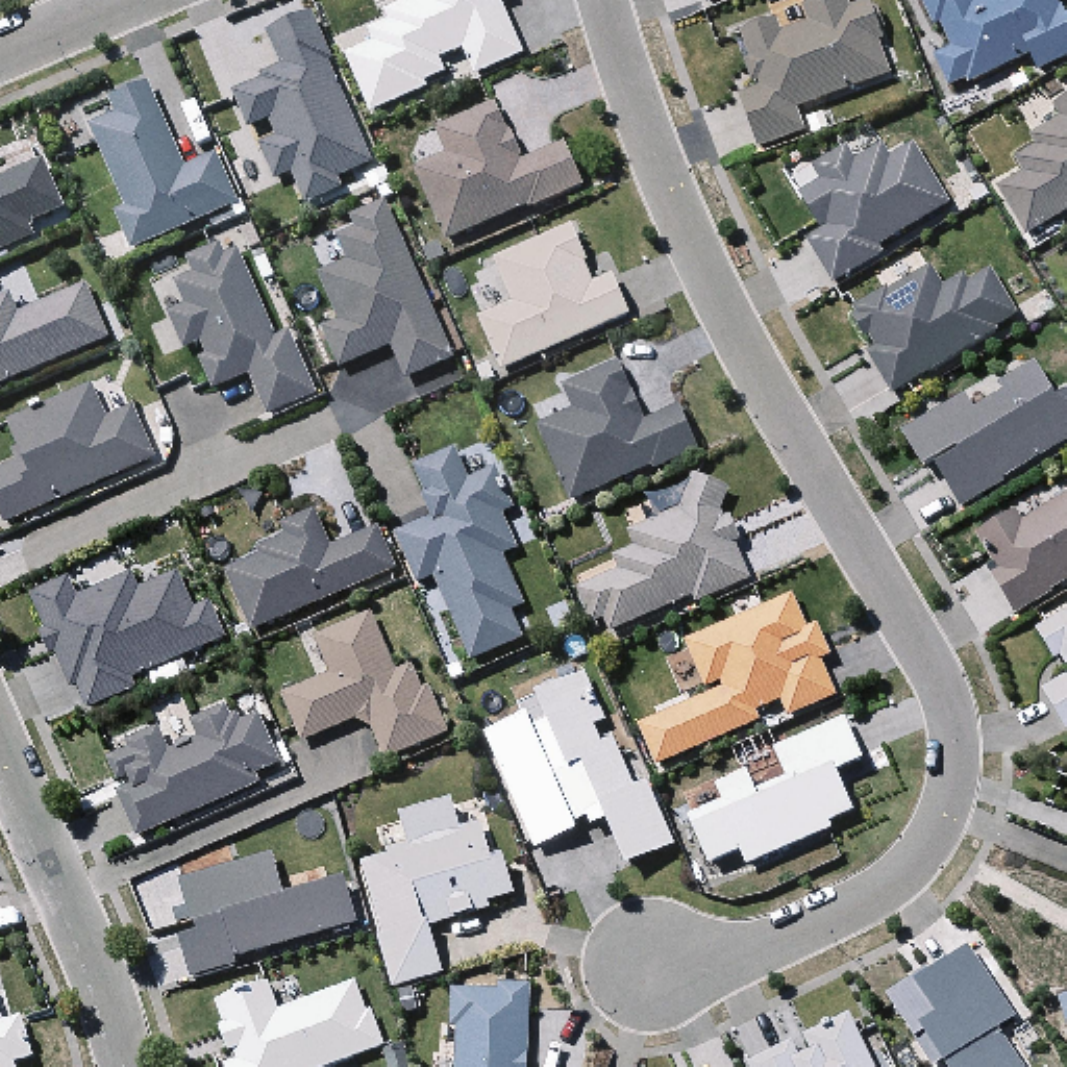} 
\end{minipage}
}
\subfigure[]{
\begin{minipage}[b]{0.25\linewidth}
\includegraphics[width=\linewidth]{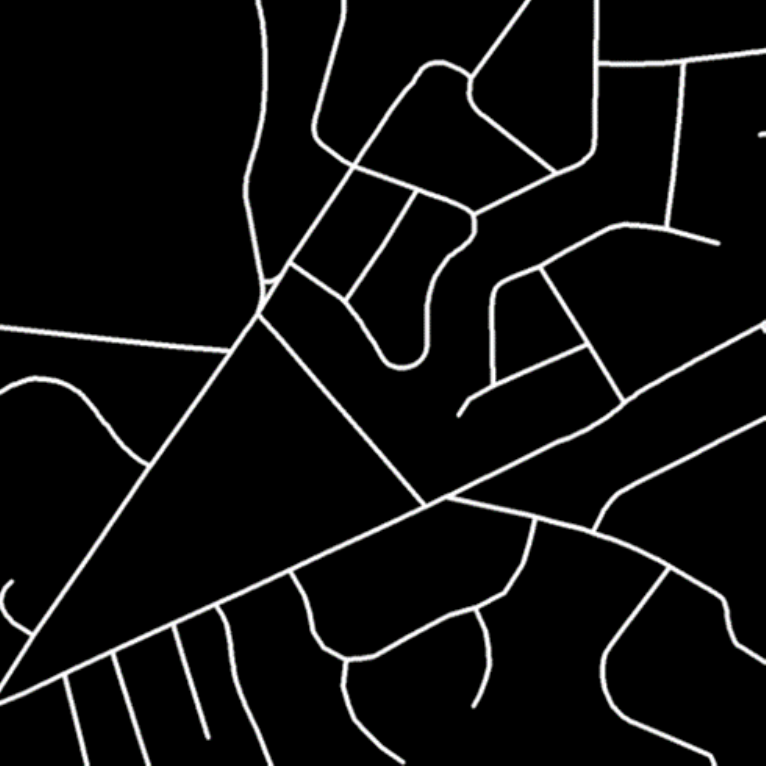} \\ \\
\includegraphics[width=\linewidth]{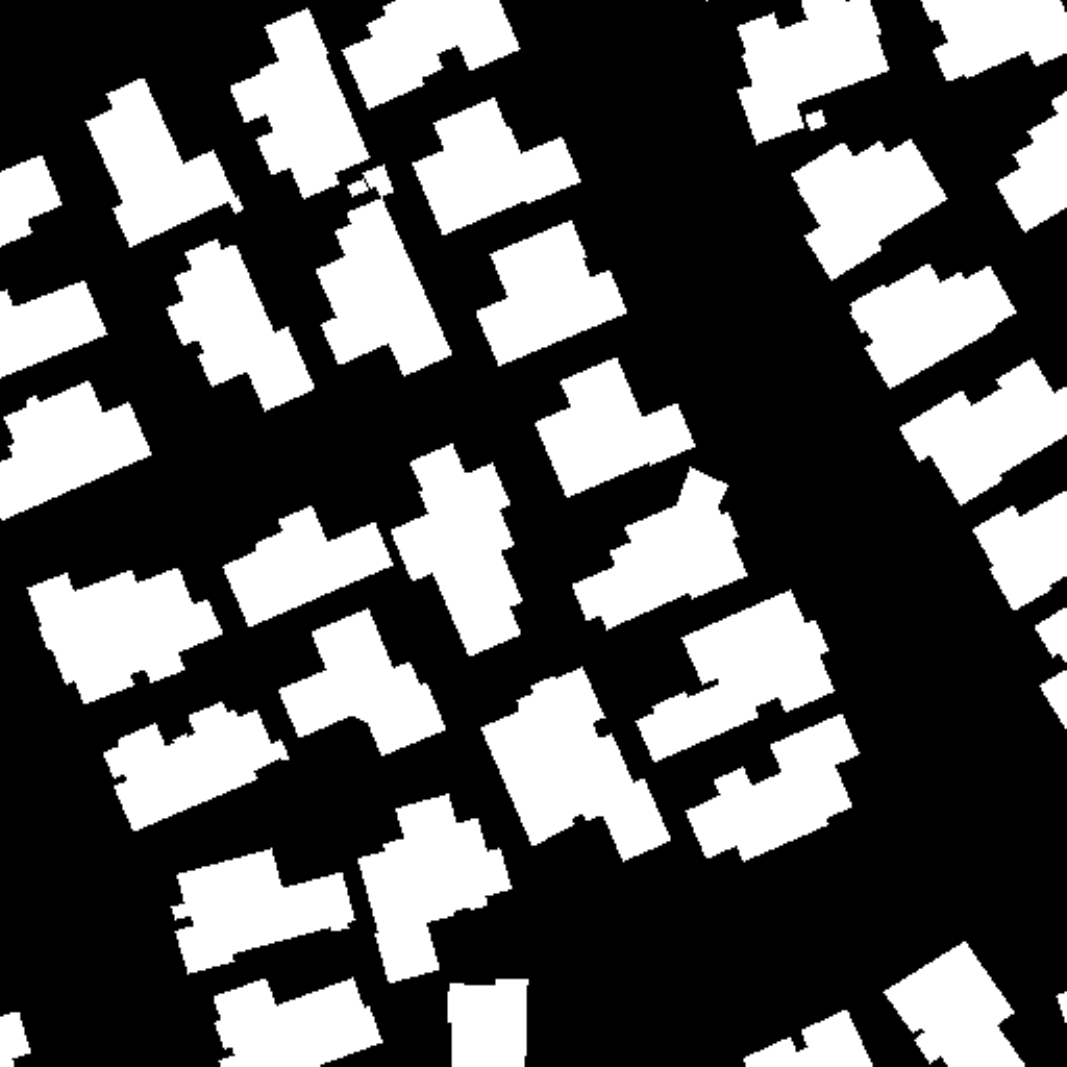}
\end{minipage}
}
\subfigure[]{
\begin{minipage}[b]{0.25\linewidth}
\includegraphics[width=\linewidth]{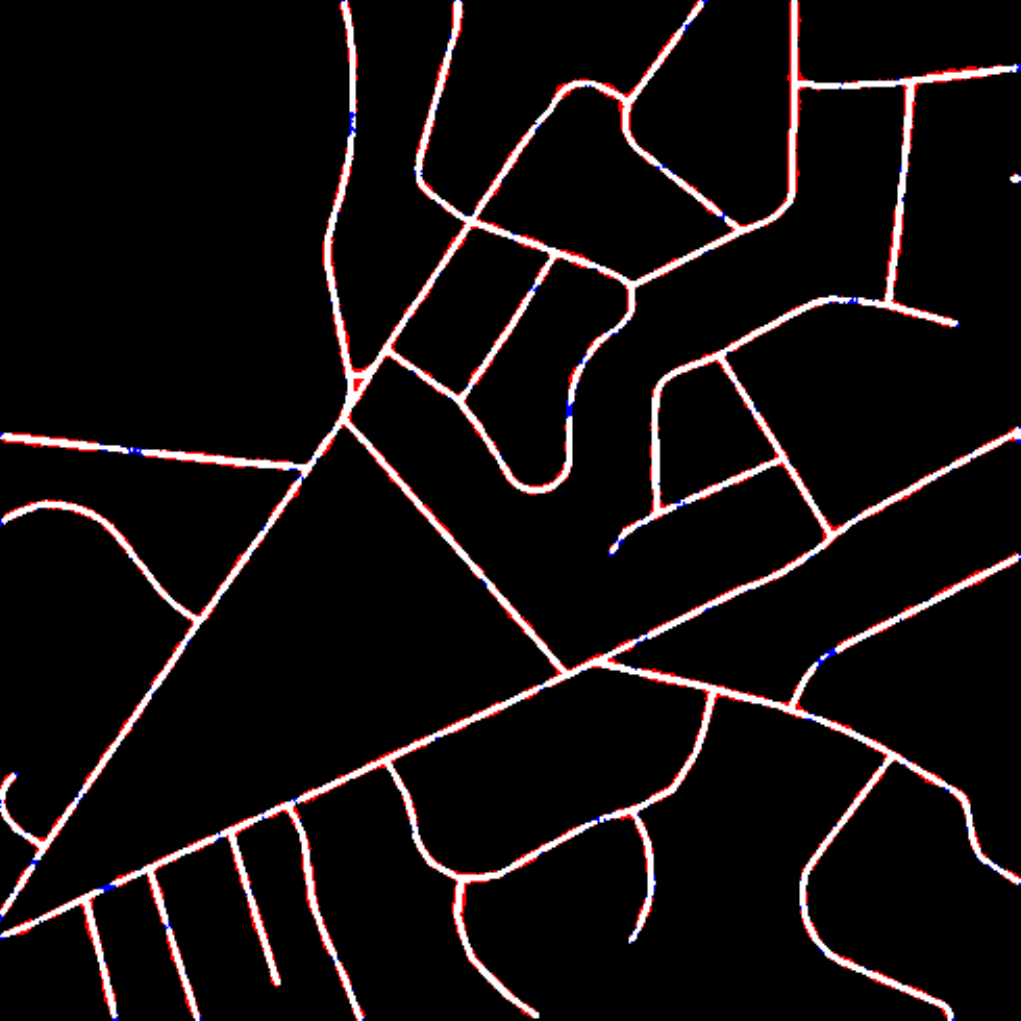} \\ \\
\includegraphics[width=\linewidth]{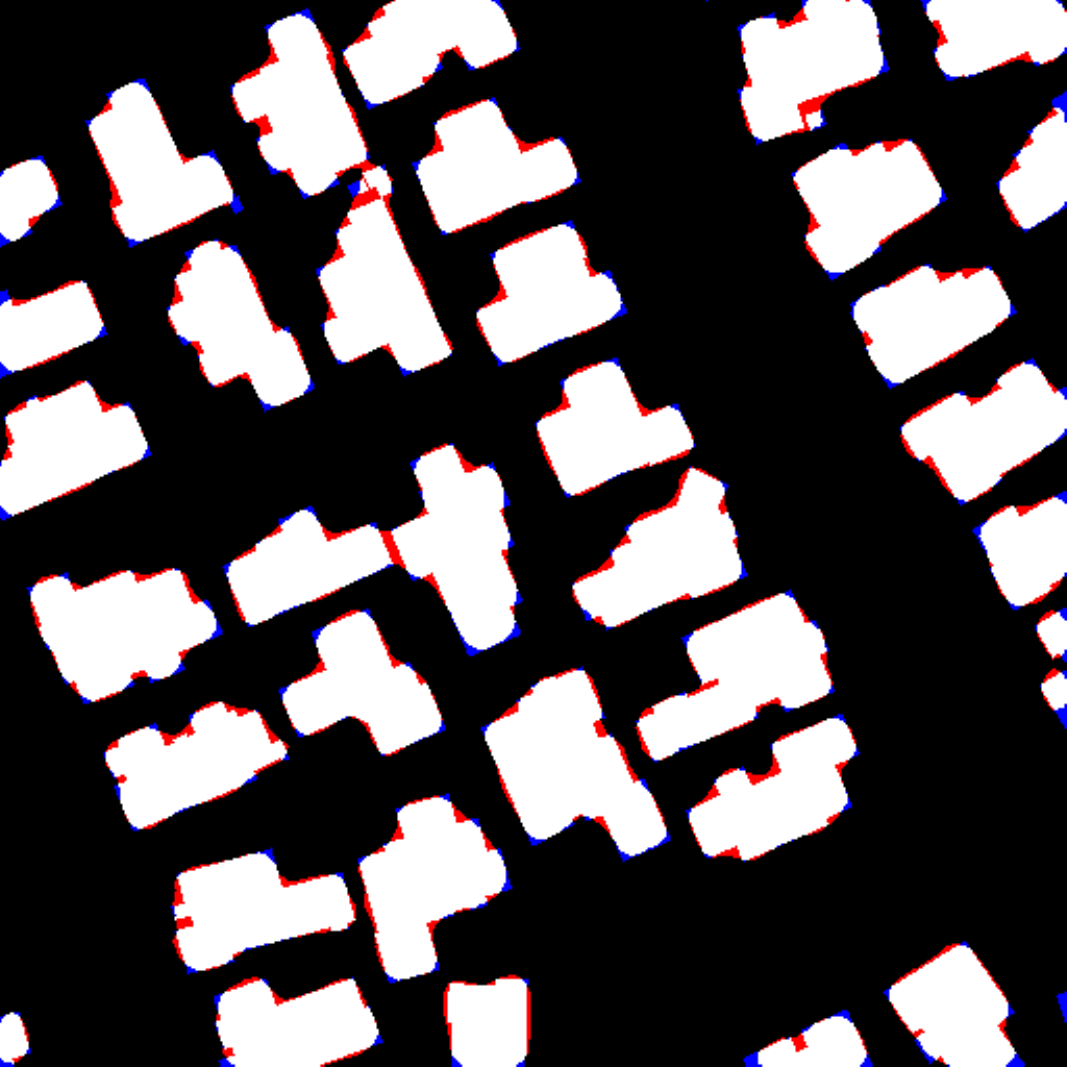}
\end{minipage}
}
\caption{Sample inference results of RSM-SS on the semantic segmentation task. The results on the Massachusetts Road and WHU datasets are shown in the first and second rows, respectively. Red areas denote false positives and blue areas denote false negatives. (a) Input image. (b) Ground truth image. (c) RSM-SS result.}
\label{Fig:seg_result}
\end{figure}

We show some inference results of the test set of the Massachusetts Road and WHU datasets in Figure \ref{Fig:seg_result}. It shows that the RSM-SS can accurately segment all the roads in the Massachusetts Road dataset and buildings in the WHU dataset. On the Massachusetts Road dataset, despite roads extending in various directions and covering extensive spatial scales, RSM-SS manages to accurately segment roads by leveraging the rich contextual information available in large VHR remote sensing images. Similarly, on the WHU dataset, despite the dense arrangement of buildings and the presence of spatial features in multiple directions, the RSM-SS is capable of extracting building features across various directions.

\subsubsection{Change Detection Task}
\label{section:4.4.2}

This subsection presents the results of comparing RSM-CD with other change detection models on the change detection task with two datasets (WHU-CD and LEVIR-CD datasets).

The accuracy comparison results on the WHU-CD dataset are shown in Table \ref{whucd_result}. RSM-CD achieves the highest IoU (0.8496) and F1-score (0.9187) on this dataset, outperforming all other change detection models. Given the very high spatial resolution of the WHU-CD dataset remote sensing images (0.075 m/pixel), a remote sensing image with a normal size (256×256 pixels) may only contain a few buildings or parts of buildings, thereby losing a significant amount of contextual information. Due to its linear complexity, RSM-CD is capable of processing large VHR remote sensing images. This allows RSM-CD to utilize the rich contextual information present in large images to accurately identify changed buildings.

\begin{table}[!ht]
\caption{Accuracy Comparison on the WHU-CD Dataset. The Best Values Are Highlighted in Bold.}
\label{whucd_result}
\centering
% \scalebox{0.8}{
\begin{tabular}{lcccc}
\toprule
Methods & P (\%) & R (\%) & F1 (\%) & IoU (\%) \\
\midrule
FC-EF~\cite{fcef} & 78.86 & 78.64 & 78.75 & 64.95 \\
FC-Siam-Diff~\cite{fcef} & 84.73 & 87.31 & 86.00 & 75.44 \\
FC-Siam-Conc~\cite{fcef} & 78.86 & 78.64 & 78.75 & 64.95\\ 
DTCDSCN~\cite{dtcdstn} & 63.92 & 82.30 & 71.95 & 56.19 \\
DSIFN~\cite{dsifn} & 91.44 & 89.75 & 90.59 & 82.79 \\
STANet~\cite{sta} & 79.37 & 85.50 & 82.32 & 69.95 \\
SNUNet~\cite{snu} & 85.60 & 81.49 & 83.49 & 71.67 \\
DASNet~\cite{dasnet} & 88.23 & 84.62 & 86.39 & 76.04 \\
HANet~\cite{hanet} & 88.30 & 88.01 & 88.16 & 78.82 \\
CDNet~\cite{cdnet} & 91.75 & 86.89 & 89.25 & 80.59 \\
DDCNN~\cite{ddcnn} & \textbf{93.71} & 89.12 & 91.36 & 84.09 \\
BIT~\cite{bit} & 86.64 & 81.48 & 83.98 & 72.39 \\
MTCNet~\cite{mtcnet} & 75.10 & \textbf{91.90} & 82.65 & 70.43 \\
MSCANet~\cite{mscanet} & 91.10 & 89.86 & 90.47 & 82.60 \\
\midrule
RSM-CD & 93.37 & 90.42 & \textbf{91.87} & \textbf{84.96} \\
\bottomrule
\end{tabular}
% }
\end{table}

The accuracy comparison results on the LEVIR-CD dataset are summarized in Table \ref{levircd_result}. RSM-CD outperforms all the other models, achieving the highest IoU (0.8366) and F1-score (0.9110) on this dataset. In the LEVIR-CD dataset, the presence of buildings with multiple orientations and arrangements in the bitemporal remote sensing images underscores the importance of extracting large features in multiple directions. The omnidirectional selective scan module of RSM-CD can extract large spatial features of buildings from various directions, thereby accurately identifying changed buildings.

\begin{table}[!ht]
\caption{Accuracy Comparison on the LEVIR-CD Dataset. The Best Values Are Highlighted in Bold.}
\label{levircd_result}
\centering
% \scalebox{0.8}{
\begin{tabular}{lcccc}
\toprule
Methods & P (\%) & R (\%) & F1 (\%) & IoU (\%) \\
\midrule
FC-EF~\cite{fcef} & 86.91 & 80.17 & 83.40 & 71.53 \\
FC-Siam-Diff~\cite{fcef} & 89.53 & 83.31 & 86.31 & 75.91 \\
FC-Siam-Conc~\cite{fcef} & 91.99 & 76.77 & 83.69 & 71.96\\ 
DTCDSCN~\cite{dtcdstn} & 88.53 & 86.83 & 87.67 & 78.05\\ 
DSIFN~\cite{dsifn} & \textbf{94.02} & 82.93 & 88.13 & 78.77 \\
STANet~\cite{sta} & 83.81 & \textbf{91.00} & 87.26 & 77.39 \\
SNUNet~\cite{snu} & 89.18 & 87.17 & 88.16 & 78.83 \\
HANet~\cite{hanet} & 91.21 & 89.36 & 90.28 & 82.27 \\
CDNet~\cite{cdnet} & 91.60 & 86.50 & 88.98 & 80.14 \\
DDCNN~\cite{ddcnn} & 91.85 & 88.69 & 90.24 & 82.22 \\
BIT~\cite{bit} & 89.24 & 89.37 & 89.30 & 80.68 \\
ChangeFormer~\cite{changeformer} & 92.05 & 88.80 & 90.40 & 82.47 \\
MTCNet~\cite{mtcnet} & 90.87 & 89.62 & 90.24 & 82.22 \\
MSCANet~\cite{mscanet} & 91.30 & 88.56 & 89.91 & 81.66 \\
AMTNet-50~\cite{amtnet} & 91.82 & 89.71 & 90.76 & 83.08 \\
\midrule
RSM-CD & 92.52 & 89.73 & \textbf{91.10} & \textbf{83.66} \\
\bottomrule
\end{tabular}
% }
\end{table}

\begin{figure}[!ht] %子图加并列
\centering
\subfigure[]{
\begin{minipage}[b]{0.213\linewidth}
\includegraphics[width=\linewidth]{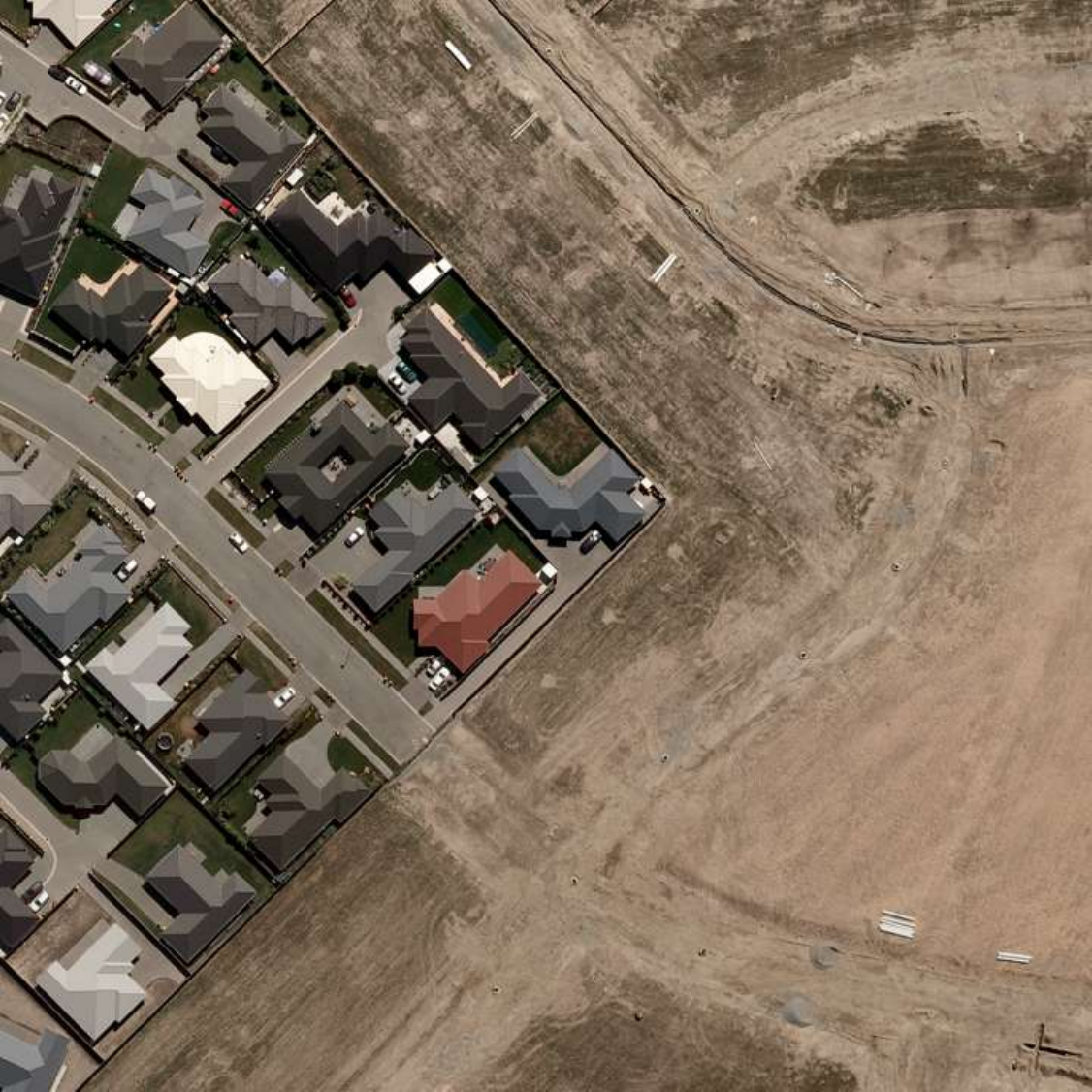} \\ \\
\includegraphics[width=\linewidth]{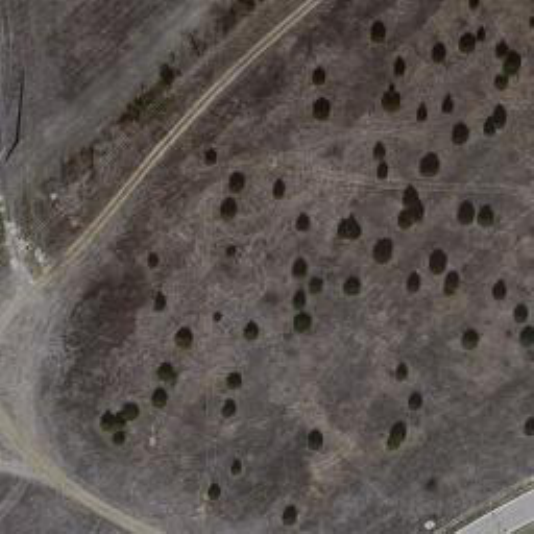}
\end{minipage}
}
\subfigure[]{
\begin{minipage}[b]{0.213\linewidth}
\includegraphics[width=\linewidth]{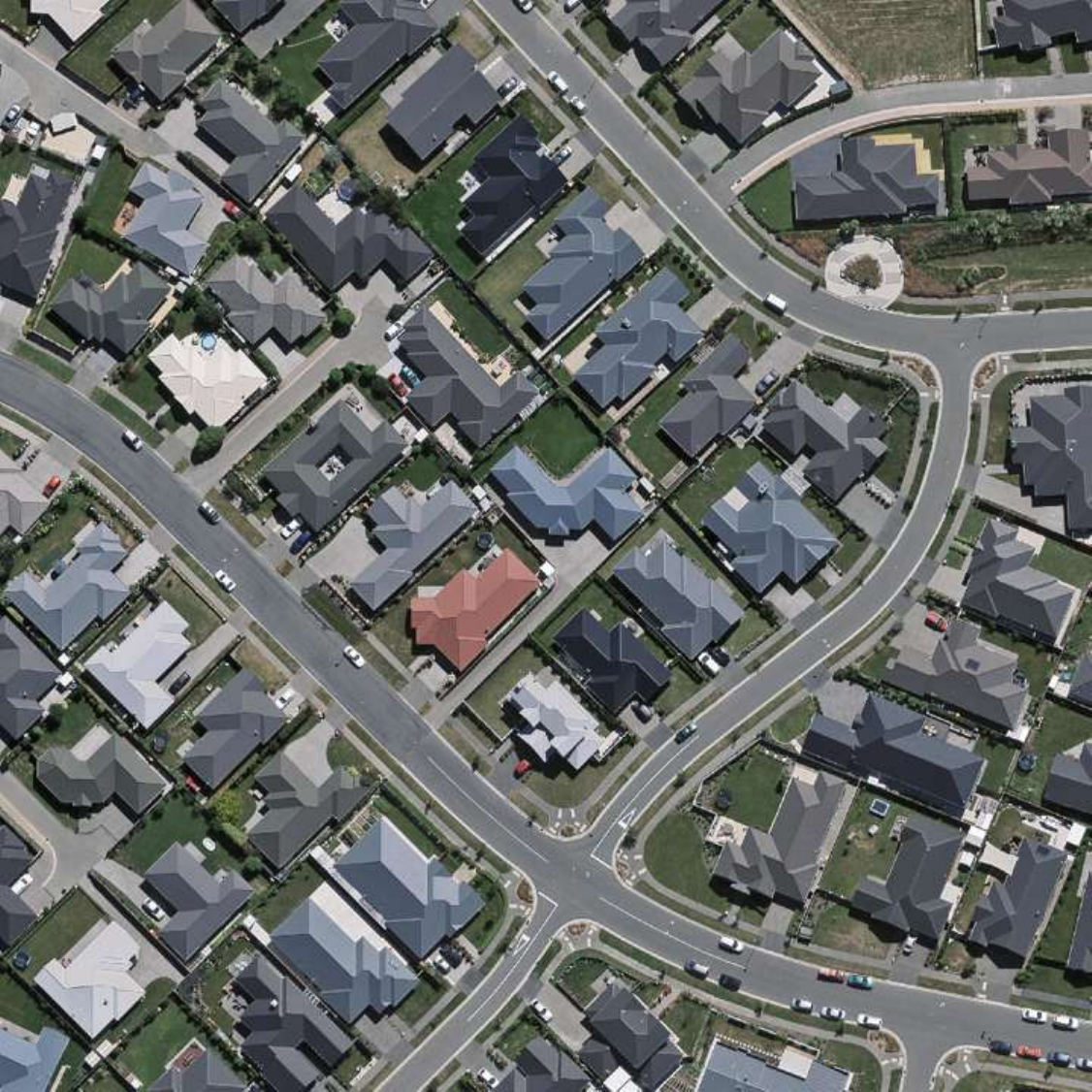} \\ \\
\includegraphics[width=\linewidth]{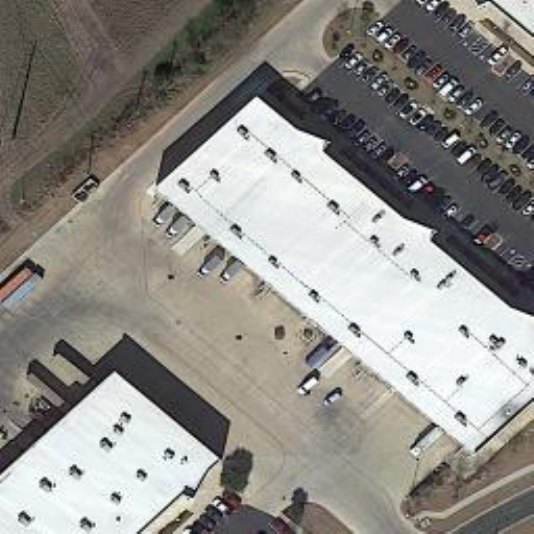}
\end{minipage}
}
\subfigure[]{
\begin{minipage}[b]{0.213\linewidth}
\includegraphics[width=\linewidth]{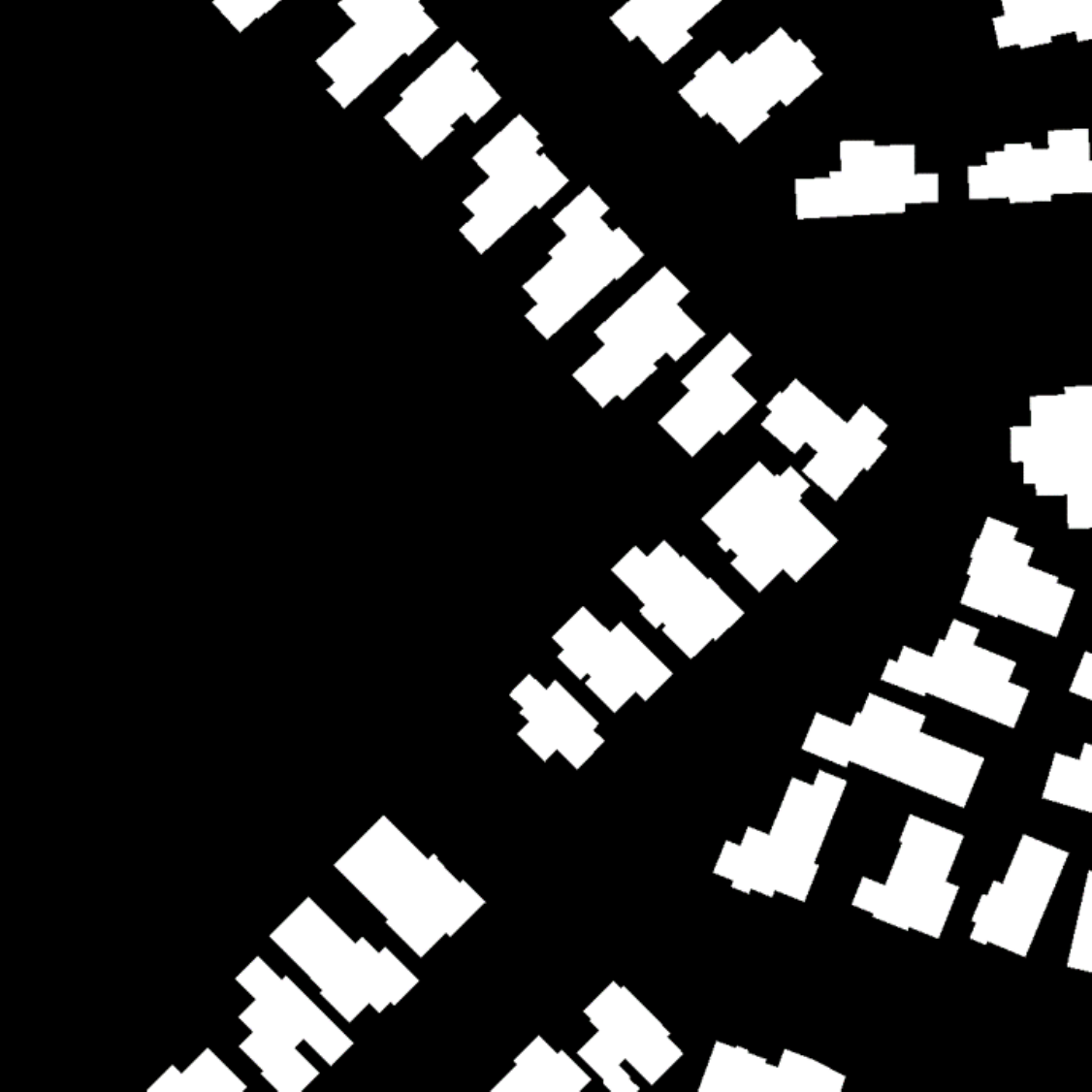} \\ \\
\includegraphics[width=\linewidth]{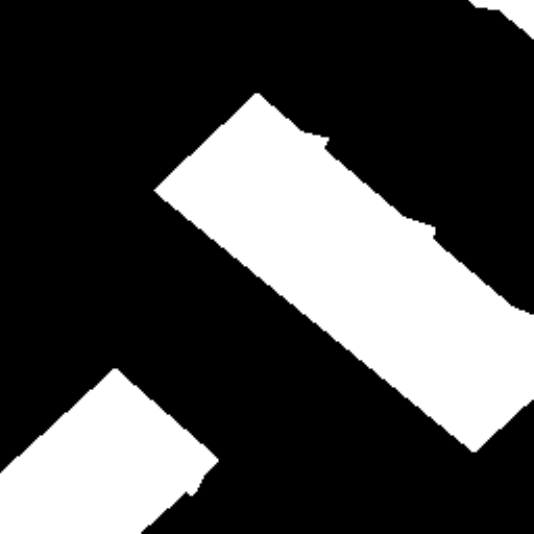}
\end{minipage}
}
\subfigure[]{
\begin{minipage}[b]{0.213\linewidth}
\includegraphics[width=\linewidth]{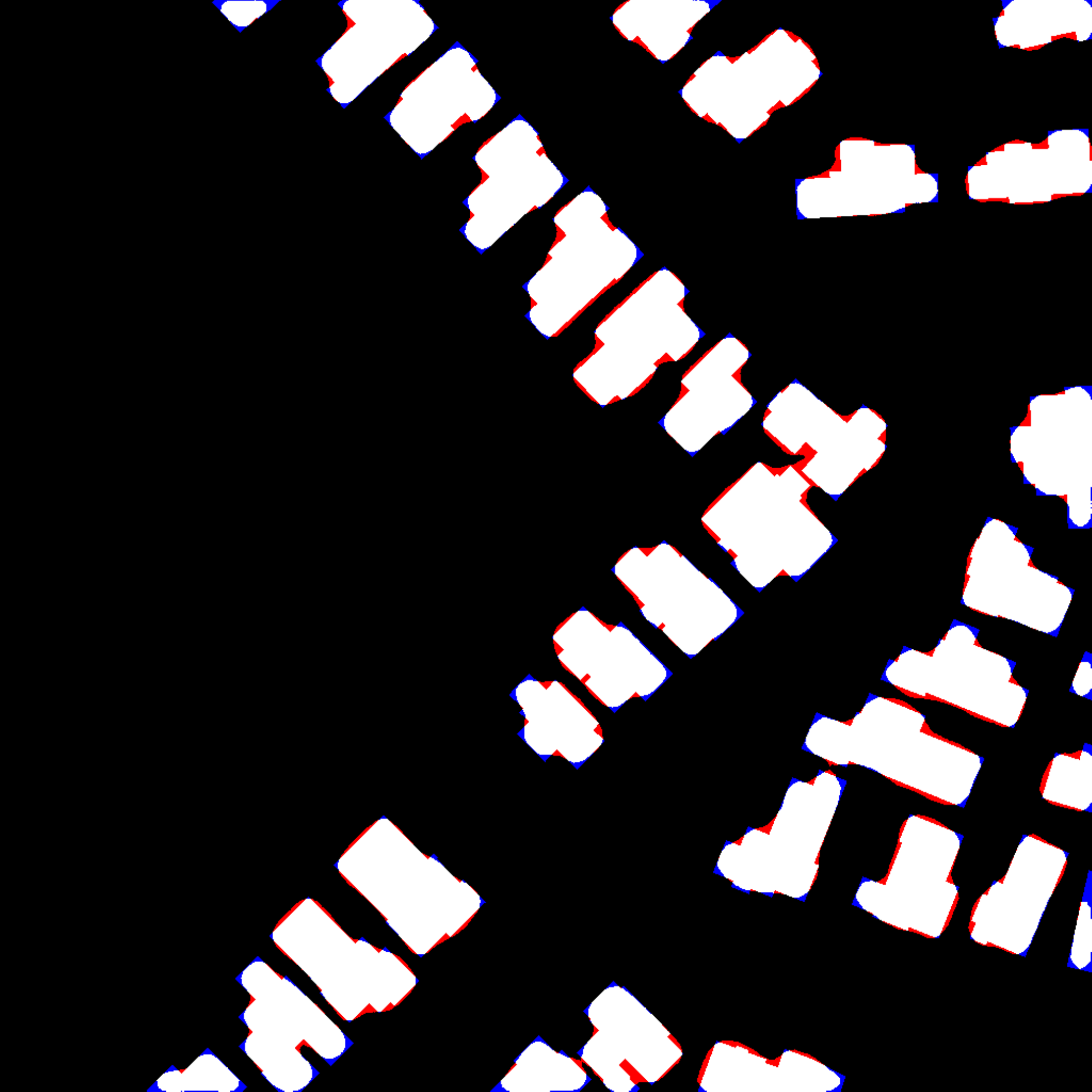} \\ \\
\includegraphics[width=\linewidth]{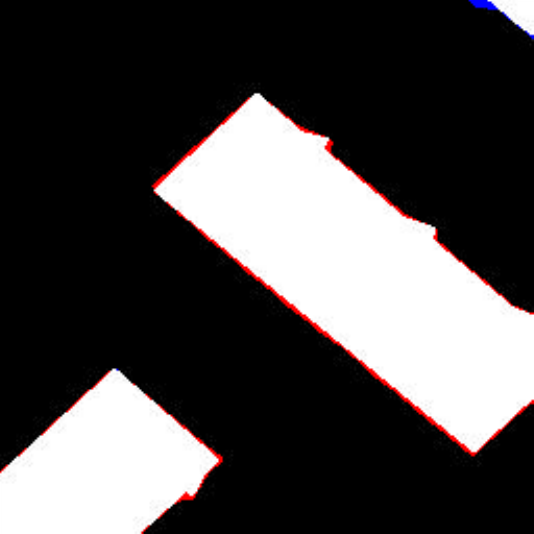}
\end{minipage}
}
\caption{Sample inference results of RSM-CD on the change detection task. The results on the WHU-CD and LEVIR-CD datasets are shown in the first and second rows, respectively. Red areas denote false positives and blue areas denote false negatives. (a) T1 image. (b) T2 image. (c) Ground truth image. (d) RSM-CD result.}
\label{Fig:sv_result}
\end{figure}

We show some inference results for the test sets of the WHU-CD and LEVIR-CD datasets in Figure \ref{Fig:sv_result}. It shows that RSM-CD can accurately detect all the changed buildings in the WHU-CD and LEVIR-CD datasets. On the WHU-CD dataset, the high spatial resolution of remote sensing images necessitates the use of large images to preserve ample contextual information. The linear complexity of RSM-CD enables it to process large VHR remote sensing images. By leveraging the rich contextual information available in the bitemporal images, RSM-CD can accurately identify changed buildings. On the LEVIR-CD dataset, where buildings exhibit edge features in multiple directions, the ability to perform selective scanning in multiple directions allows RSM-CD to extract building features from various orientations, accurately identifying changed buildings.

\subsection{Impact of Image Size and Spatial Resolution}

The extensive contextual information and high-resolution spatial features of VHR remote sensing images are crucial for dense prediction tasks. To investigate the impact of contextual information and spatial features on dense prediction tasks in VHR remote sensing, we experimented with semantic segmentation and change detection tasks using images of varying sizes and downsampling factors, as cropping images into small patches would lose contextual information and downsampling images would lose spatial features.

\begin{figure}[!ht]
	\centering
		\includegraphics[width=\linewidth]{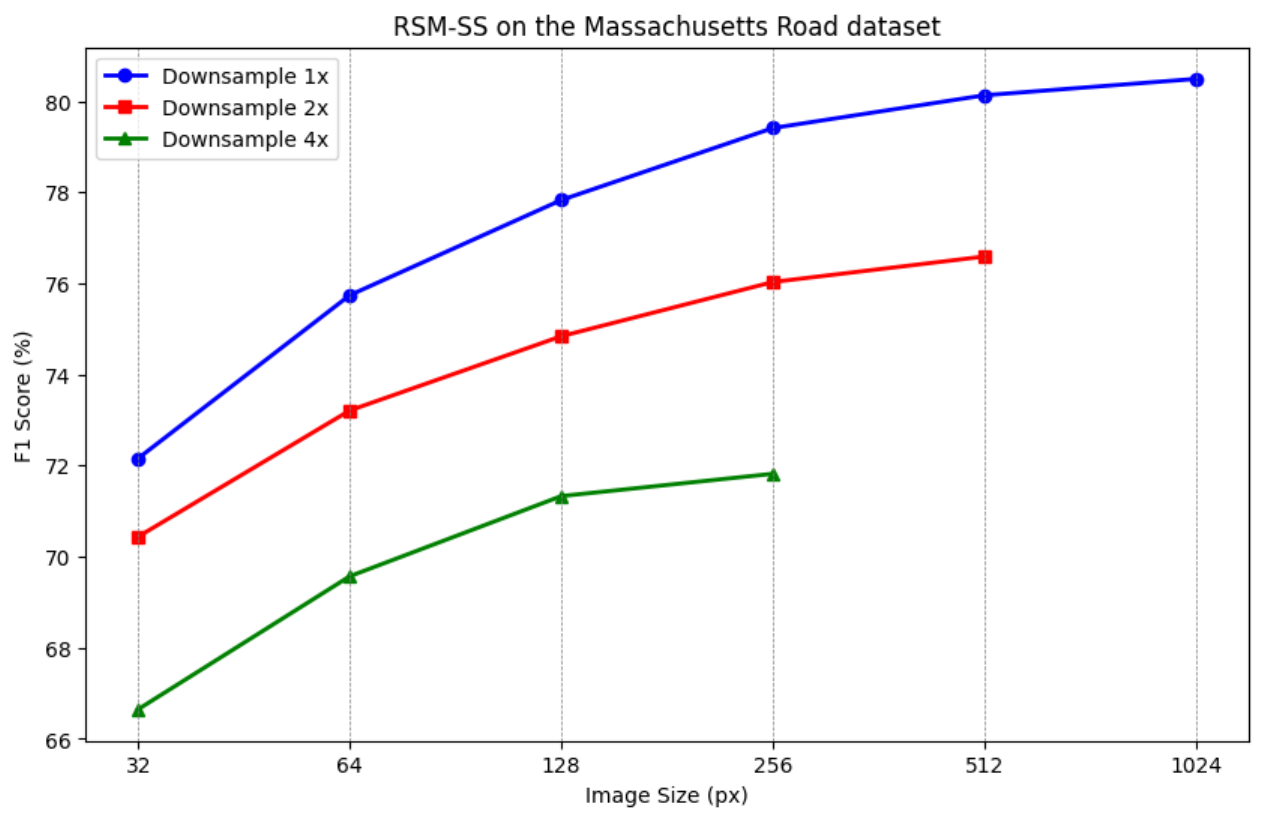}
	\caption{Performance of RSM-SS on the Massachusetts Roads dataset with different image sizes and downsampling ratios.}
 \label{Fig:discuss_mszs}
\end{figure}

In our semantic segmentation experiments on the Massachusetts Roads dataset, where roads are the objects of interest, we first downsampled the remote sensing images by factors of 1 (no downsampling), 2, and 4. Then, we cropped the images to sizes of 32, 64, 128, 256, 512, and 1024 pixels. It is important to note that images downsampled by a factor of 2 have a maximum size of 512 pixels, and those downsampled by a factor of 4 have a maximum size of 256 pixels. For two images of the same size but different downsampling ratios (ratio 1, ratio 2), the latter has a ratio2/ratio1 times the spatial range of the former, which means that an image with a size of 512 and a downsampling ratio of 1 has the same spatial range as an image with a size of 256 and a downsampling ratio of 2. We used the F1-score as a metric to evaluate the model's performance across different image sizes and downsampling ratios, and the results are illustrated in Figure \ref{Fig:discuss_mszs}.

The results indicate that the model's performance improves with increasing image size, regardless of the downsampling ratio. For images of the same size, those with a higher downsampling ratio perform worse, despite having more contextual information. This could be attributed to the elongated nature of roads, which extend in various directions across the image. Downsampling the images results in a significant loss of road spatial features, making it difficult to segment roads. Cropping the images into smaller patches leads to a substantial loss of contextual information, which hampers the ability to determine the roads' extension directions and segment them effectively. Thus, both large amounts of contextual information and high-resolution spatial features are important for segmenting roads.

\begin{figure}[!ht]
	\centering
		\includegraphics[width=\linewidth]{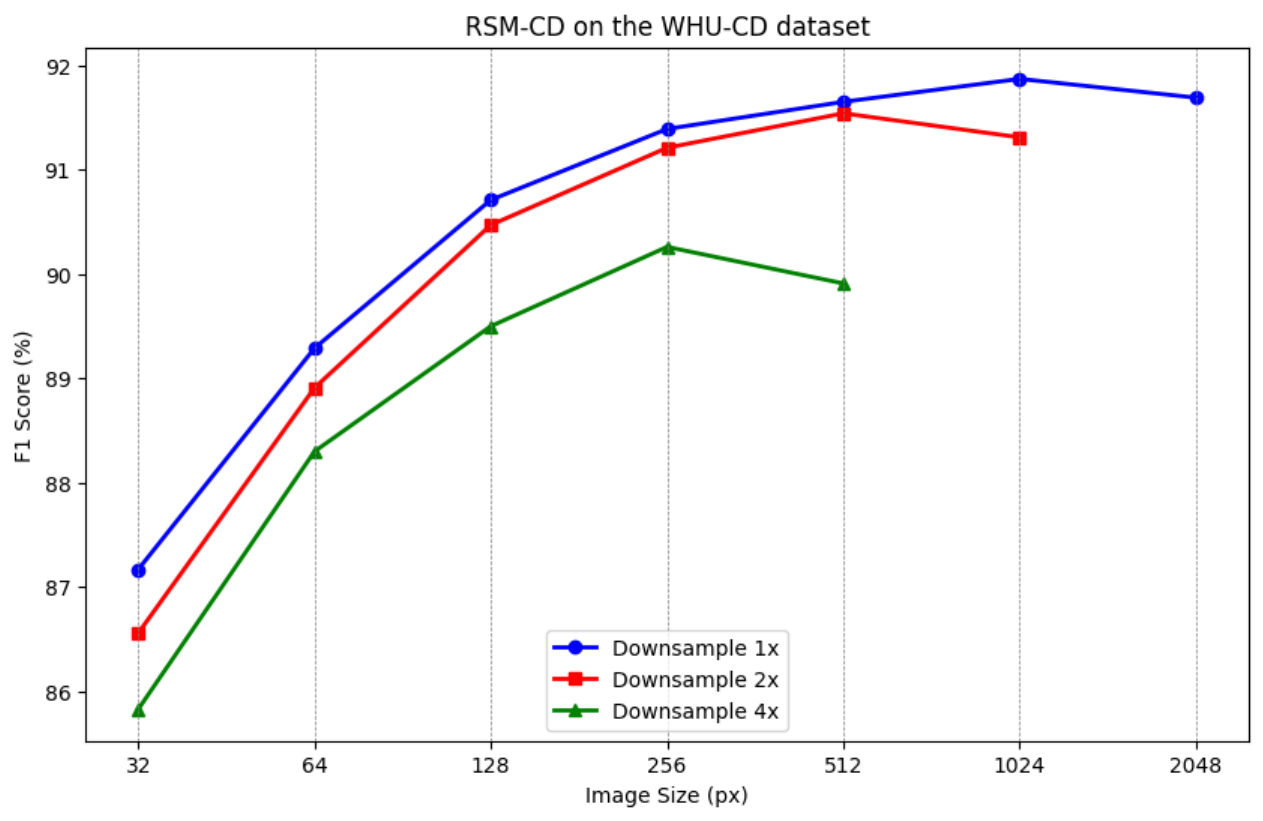}
	\caption{Performance of RSM-CD on the WHU-CD dataset with different image sizes and downsampling ratios.}
 \label{Fig:discuss_whu}
\end{figure}

In the change detection task, we conducted experiments on the WHU-CD dataset, where the changed objects are buildings. The strategies for image downsampling and cropping were the same as those applied to the Massachusetts Roads dataset, with the F1-score serving as the evaluation metric. The performance of the model across different image sizes and downsampling ratios is illustrated in Figure \ref{Fig:discuss_whu}. The results show that the model's performance initially increases with the size of the image, reaching a peak before starting to decline. The image size peaks for downsampling ratios of 1, 2, and 4 correspond to 256, 512, and 1024, respectively, each offering the same level of spatial range. Furthermore, at any given size, the model performs worse on images with a higher downsampling ratio. This may be due to the presence of spatial features within individual buildings and between multiple buildings, necessitating a certain level of contextual information and high-resolution spatial features to accurately identify changed buildings. Downsampling results in a significant loss of spatial features, substantially reducing the model's performance. Cropping images to a certain size makes larger patches containing more contextual information, which benefits the model in detecting changed buildings. However, when there is too much contextual information, including a lot of irrelevant details for identifying a specific changed building, the model's performance decreases.

The experiments on the Massachusetts Roads and WHU-CD datasets highlight the importance of contextual information and high-resolution spatial features for dense prediction tasks in VHR remote sensing, which aligns with the experimental results from FCCDN~\cite{fccdn} and SwinB-CNN~\cite{hybrid1}. The model's performance varies in response to the loss of contextual information and spatial features, with a more significant decrease in performance when downsampling road images than when downsampling building images. A common finding is that the model performs better on higher spatial resolution images. Moreover, the higher the spatial resolution of the images is, the larger the image size required to contain the same level of contextual information, thus larger images are needed for the model to perform optimally. Therefore, VHR remote sensing images and models capable of processing large images are crucial for dense prediction tasks in remote sensing. The linear complexity of the RSM enables it to handle large VHR remote sensing images, thereby achieving excellent results in dense prediction tasks for VHR remote sensing.

\subsection{Handling Large Remote Sensing Images}

Large remote sensing images contain rich contextual information, which is crucial for dense prediction tasks. To demonstrate the superiority of the RSM over CNN-based and transformer-based models in processing large remote sensing images, we conducted comparative experiments on the WHU-CD dataset for change detection tasks. Given that the original WHU-CD data are completely large remote sensing images, we can crop them into image patches of various sizes for our experiments. We compared RSM-CD with ResNet-50-CD and ViT-S-CD, which have a similar number of parameters. All these models have the same framework, with the only difference lying in the encoder. ResNet-50-CD replaces the encoder of RSM-CD with ResNet-50, and ViT-S-CD replaces it with ViT-S. 

The comparison results are shown in Table \ref{flops_comparsion}, where the batch size is set to 1 when calculating the FLOPs of the models. OOM stands for out of memory, indicating that the model training could not proceed because the required memory exceeded the machine's available memory. As the image size increased, the computational costs of ResNet-50-CD and RSM-CD exhibited a linear growth trend because they have linear complexity. In contrast, as ViT-S-CD has quadratic complexity, it exhibited a quadratic growth trend and significantly exceeded the computational costs of ResNet-50-CD and RSM-CD on large remote sensing images. This complexity led to the inability of the model to train on images of sizes 1024 and 2048 due to excessive memory requirements. 

In terms of model performance, RSM-CD achieved the best results on image size 1024 with a similar parameter count and lower computational cost, outperforming ResNet-50-CD and ViT-S-CD of all valid image sizes. Although ResNet-50-CD has linear complexity, it cannot globally model the context of remote sensing images and only has a local effective receptive field, which fails to utilize the rich contextual information of large remote sensing images effectively. While ViT-S-CD can globally model the context of remote sensing images, its quadratic complexity results in a rapid increase in computational cost and memory usage, restricting it to processing only small image patches. These small patches contain very limited contextual information, disadvantaging ViT-S-CD in change detection tasks. RSM-CD has linear complexity and can model the context of remote sensing images globally across multiple directions, extracting large spatial features from various directions, thus achieving superior results in change detection tasks.

\begin{table}[!ht]
\caption{Comparative experiments on the Computational cost and model performance, which were conducted on the WHU-CD dataset.}
\label{flops_comparsion}
\centering
\begin{tabular}{lcccc} 
\toprule
Model & Param (M) & Image size & GFLOPs & F1 (\%) \\
\midrule
ResNet-50-CD & 30.4 & 32 & 1.7 & 86.42 \\
ViT-S-CD & 26.9 & 32 & 1.3 & 86.74 \\
RSM-CD & 27.9 & 32 & 2.9 & 87.17 \\
\midrule
ResNet-50-CD & 30.4 & 64 & 3.9 & 87.33 \\
ViT-S-CD & 26.9 & 64 & 3.6 & 88.53 \\
RSM-CD & 27.9 & 64 & 4.1 & 89.17 \\
\midrule
ResNet-50-CD & 30.4 & 128 & 6.8 & 88.07 \\
ViT-S-CD & 26.9 & 128 & 7.6 & 89.74 \\
RSM-CD & 27.9 & 128 & 7.2 & 90.71 \\
\midrule
ResNet-50-CD & 30.4 & 256 & 13.5 & 88.41 \\
ViT-S-CD & 26.9 & 256 & 16.8 & 90.33 \\
RSM-CD & 27.9 & 256 & 15.7 & 91.39 \\
\midrule
ResNet-50-CD & 30.4 & 512 & 53.7 & 88.53 \\
ViT-S-CD & 26.9 & 512 & 107.9 & 90.52 \\
RSM-CD & 27.9 & 512 & 50.2 & 91.65 \\
\midrule
ResNet-50-CD & 30.4 & 1024 & 194.7 & 88.37 \\
ViT-S-CD & 26.9 & 1024 & 973.4 & OOM \\
RSM-CD & 27.9 & 1024 & 185.2 & 91.87 \\
\midrule
ResNet-50-CD & 30.4 & 2048 & 793.9 & 88.45 \\
ViT-S-CD & 26.9 & 2048 & 9148.7 & OOM \\
RSM-CD & 27.9 & 2048 & 742.8 & 91.49 \\

\bottomrule
\end{tabular}
\end{table}

% \section{Analysis Experiments}

\section{Discussion}

% \subsection{Expectations and Limitations}

In the realm of VHR remote sensing, contextual information within images is crucial for dense prediction tasks. However, current models based on CNNs and transformers struggle to process large VHR remote sensing images effectively. CNN-based models, limited by their local convolution operations, fail to model the global contextual information of VHR remote sensing images. Transformer-based models, due to their quadratic complexity, are incapable of handling large VHR images. These models typically resort to processing smaller image patches, which contain limited contextual information, thus hindering their performance on dense prediction tasks.

To address these issues, we propose the RSM for dense prediction tasks in VHR remote sensing. The RSM, characterized by its linear complexity and global modeling capabilities, can process large VHR remote sensing images and model their global contextual information. It is capable of extracting large spatial features in multiple directions, thus effectively facilitating dense prediction tasks. The experimental results of the semantic segmentation and change detection tasks demonstrate that, despite RSM-SS and RSM-CD employing simple model architectures without any sophisticated modules or training techniques, they achieve state-of-the-art performance in their respective tasks. This validates the potential of the RSM in dense prediction tasks for VHR remote sensing. We hope that the RSM can serve as a baseline in the field, promoting the development of SSM-based methods within VHR remote sensing.

% Despite the notable performance of the RSM in dense prediction tasks for VHR remote sensing, it also exhibits some limitations. On the one hand, the models for semantic segmentation and change detection tasks are overly simplistic, not fully leveraging the potential of SSM. On the other hand, dense prediction tasks in VHR remote sensing require extensive training data, thus limiting RSM’s applicability in tasks with limited labeled data.

Despite the notable performance of RSM in dense prediction tasks for ultra-high-resolution remote sensing, it also exhibits some limitations. On one hand, the models for semantic segmentation and change detection tasks are overly simplistic, not fully leveraging the potential of SSMs. The experimental results of RSM-CD on the WHU-CD dataset indicate optimal performance when the image size is 1024. However, a reduction in model performance is observed as the image size increases to 2048. This decline could be attributed to the inherent simplicity of the RSM-CD, which may limit its capability to effectively process the global contextual information contained within large remote sensing images. On the other hand, dense prediction tasks in ultra-high-resolution remote sensing require extensive training data, thus limiting RSM’s applicability in tasks lacking labeled data. 

In the future, we will develop more complex and effective SSM-based models to enhance their capacity for processing global contextual information in large remote sensing images, further exploring the potential of SSM-based models in remote sensing dense prediction tasks. At the same time, we will investigate the potential of SSM-based models in remote sensing self-supervised learning, enabling the use of vast quantities of unlabeled remote sensing images for self-supervised training of SSM-based models, and making these models applicable to tasks and scenarios where labeled data are scarce or unavailable.

\section{Conclusion}

We have proposed the Remote Sensing Mamba for dense prediction tasks in VHR remote sensing imagery, addressing the limitations of CNN-based models in global context information modeling and the challenges of transformer-based models handling large remote sensing images. Our model can process large VHR remote sensing images with rich contextual information with linear complexity. Through selective scanning in multiple directions, the RSM models global context information and extracts large spatial features across various directions, thereby efficiently accomplishing dense prediction tasks.

Experiments on semantic segmentation and change detection tasks demonstrate the superior performance of the RSM across different objects. Leveraging the State Space Model for processing large images and modeling global context information, the RSM operates on VHR remote sensing images without the need to segment these images into smaller patches, which is achieved through its linear complexity. By modeling globally in various directions, the RSM captures large spatial features from multiple perspectives, leading to outstanding performance in dense prediction tasks. We envision RSM to serve as a baseline for dense prediction tasks in VHR remote sensing, promoting further development of SSM-based approaches in this field.

% % use section* for acknowledgment
\section*{Acknowledgment}
This work was done during the internship of Sijie Zhao at Shanghai Artificial Intelligence Laboratory. 
% This work is partially supported by the National Key R\&D Program of China(NO.2022ZD0160101), in part by the National Natural Science Foundation of China under Grant 42071297, in part by the AI \& AI for Science Project of Nanjing University under Grant 02091480605203, and in part by the Fundamental Research Funds for the Central Universities under Grant 020914380119. They would also like to thank the editor and the anonymous reviewers for their constructive comments.

\ifCLASSOPTIONcaptionsoff
 \newpage
\fi

\bibliographystyle{IEEEtranN}
\bibliography{paper.bib}

% that's all folks
\end{document}